\documentclass[journal]{IEEEtran}
\usepackage{graphicx}
\usepackage{amsmath}
\usepackage{amsfonts}
\usepackage{algorithmic}
\usepackage[ruled,vlined]{algorithm2e}
\usepackage{amsthm}
\usepackage{amssymb}
\newtheorem{theorem}{Theorem}[section]

\usepackage{multirow}
\usepackage{xcolor, colortbl}
\definecolor{lavender}{rgb}{0.9, 0.9, 0.98}
\usepackage[normalem]{ulem}
\usepackage{adjustbox}
\usepackage{rotating}
\usepackage{romannum}

\usepackage{bbding}
\usepackage{pifont}
\usepackage{soul}
\setstcolor{red}
\usepackage{textcomp}
\usepackage{comment}

\DeclareMathOperator*{\argmaxA}{arg\,max}
\DeclareMathOperator{\Ker}{Ker}

\ifCLASSINFOpdf
\else
\fi
\begin{document}
%
\title{LSDM: Long-Short Diffeomorphic Motion for Weakly-Supervised Ultrasound Landmark Tracking}
%
%
%

\author{Zhihua Liu,
        Bin Yang,
        Yan Shen,
        Xuejun Ni,
        Huiyu Zhou
\thanks{Zhihua Liu and Huiyu Zhou are with School of Computing and Mathematical Sciences, University of Leicester, Leicester LE1 7RH, U.K.}
\thanks{Bin Yang is with Department of Cardiovascular Sciences, College of Life Sciences, University of Leicester, University Hospitals of Leicester NHS Trust, Leicester LE1 9HN, U.K.; Nantong-Leicester Joint Institute of Kidney Science, Department of Nephrology, Affiliated Hospital of Nantong University, Nantong, 226001, China}
\thanks{Yan Shen, Associated Chief physician, Department of Emergency Medicine, Affiliated Hospital of Nantong University, No.20 Xisi Road, Nantong City, Jiangsu Province, 226001, China}
\thanks{Xuejun Ni, Chief physician, Department of Medical Ultrasound, Affiliated Hospital of Nantong University, No.20 Xisi Road, Nantong City, Jiangsu Province, 226001, China}
\thanks{Corresponding author: Huiyu Zhou. Email: hz143@leicester.ac.uk}}

%
%

\markboth{Journal of \LaTeX\ Class Files,~Vol.~14, No.~8, August~2015}%
{Shell \MakeLowercase{\textit{et al.}}: Bare Demo of IEEEtran.cls for IEEE Journals}
%



\maketitle

\begin{abstract}
Accurate tracking of an anatomical landmark over time has been of high interests for disease assessment such as minimally invasive surgery and tumor radiation therapy. Ultrasound imaging is a promising modality benefiting from low-cost and real-time acquisition. However, generating a precise landmark tracklet is very challenging, as attempts can be easily distorted by different interference such as landmark deformation, visual ambiguity and partial observation. In this paper, we propose a long-short diffeomorphic motion network, which is a multi-task framework with a learnable deformation prior to search for the plausible deformation of landmark. Specifically, we design a novel diffeomorphism representation in both long and short temporal domains for delineating motion margins and reducing long-term cumulative tracking errors. To further mitigate local anatomical ambiguity, we propose an expectation maximisation motion alignment module to iteratively optimize both long and short deformation, aligning to the same directional and spatial representation. The proposed multi-task system can be trained in a weakly-supervised manner, which only requires few landmark annotations for tracking and zero annotation for long-short deformation learning. We conduct extensive experiments on two ultrasound landmark tracking datasets. Experimental results show that our proposed method can achieve better or competitive landmark tracking performance compared with other state-of-the-art tracking methods, with a strong generalization capability across different scanner types and different ultrasound modalities.
\end{abstract}



\section{Introduction}
%
%
%
%
\IEEEPARstart{A}{ccurate} anatomical landmark tracking has attracted significant attention in various aspects within clinical workflows, especially in high-intensity modulated imaging and image-guided radiation therapy (RT) \cite{verellen2007innovations}, \cite{de2018evaluation}. Real-time accurate anatomical landmark tracking delivers precise landmark localization and movement estimation information in temporal-spatial domains, which provides clinicians a measurable therapy margin around clinical and surgical targets to increase the chance of tumor control \cite{keall2006management}. Among various imaging modalities, ultrasound is one of the most desirable technique benefited from low-cost, non-invasive, and real-time acquisition \cite{litjens2017survey}. Different from high temporal-spatial resolution images such as magnetic resonance (MR) or computed tomography (CT), ultrasound suffers from low signal-to-noise ratio, speckle decorrelation, spatial ambiguities and aliasing, making the small anatomical structures (such as tumor boundary, vessel wall) hard to be distinguished and tracked from surroundings \cite{queiros2018mitt}. 

\begin{figure}
    \centering
    \includegraphics[width=0.48\textwidth]{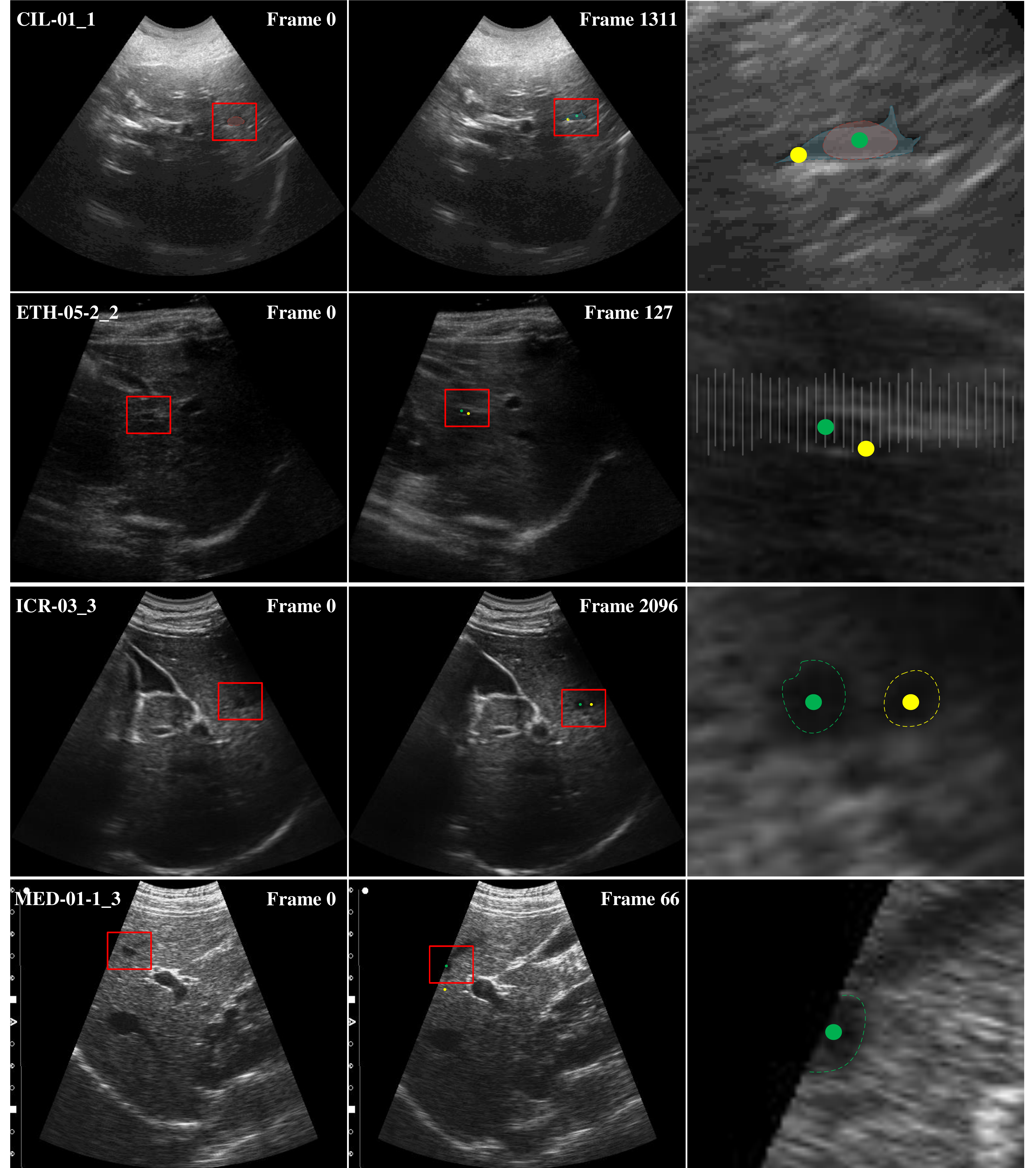}
    \caption{Illustration of ultrasound medical landmark under different challenges. From top to bottom: tracking under landmark deformation, acquisition noise, visual ambiguity and partial observation. From left to right: reference frame, example target frame, cropped region with LSDM (green) and baseline SiamFC (yellow) tracking results.}
\label{fig:1}
\end{figure}

\begin{figure}
    \centering
    \includegraphics[width=0.5\textwidth, keepaspectratio]{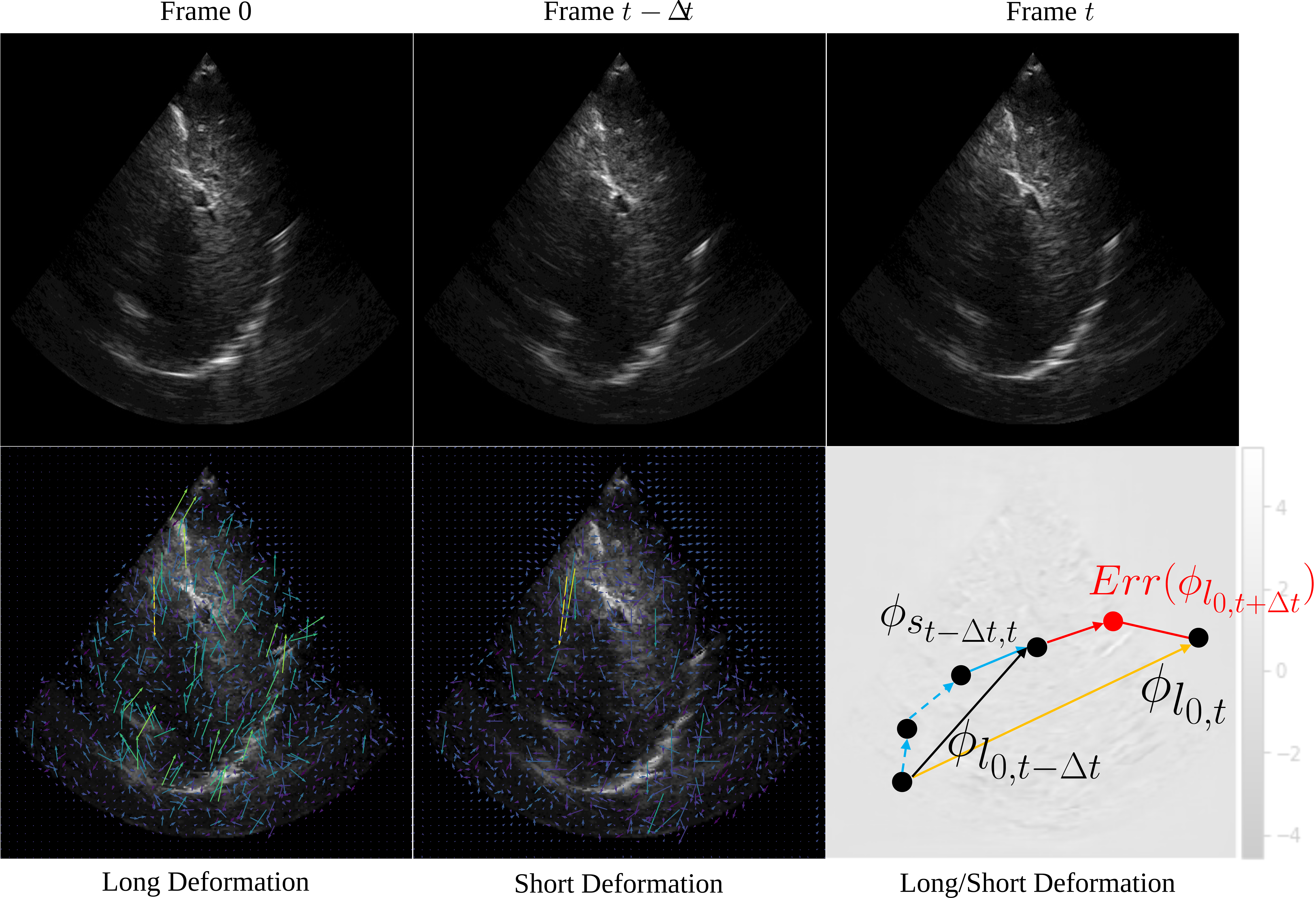}
    \caption{Illustration of Long-Short Diffeomorphic Motion. Top row denotes an example of the input image set. From top-left to top-right: Frame 0 serves as the fixed template for long range deformation. Frame $t-\Delta t$ serves the moving target for long range deformation, also adaptive template for short range deformation. Frame $t$ serves the short range target. From bottom-left to bottom right: Long range diffeomorphism vector; Short range diffeomorphism vector; Illustration of joint long-short diffeomorphism motion (blue: short range deformation, yellow: long range deformation; red: cumulative error from single range deformation).}
\label{fig:longshort}
\end{figure}

Extensive research approaches have been proposed in past decades to track anatomical landmarks. One implementation is using invasive artificial fiducial markers, which is limited with regards to surgery implementation requirements and marker migration \cite{jayarathne2018robust}, \cite{van2013critical}. Automated learning algorithms have been investigated and achieved significant improvements \cite{cifor2013hybrid}, \cite{royer2017real}. However, they cannot effectively handle the ultrasound sequences of poor quality, where the intensity is anisotropic in the temporal dimension, caused by sonar noise and shadowing effects. Moreover, most existing tracking methods can not measure the topology changes of targets between frames, (e.g. disconnected boundaries shown in Fig. \ref{fig:1}), which varies from patient to patient involving motions from respiratory, cardiac or body movement. These internal and external noise greatly influence the tracking performance, especially trackers rely on learning similarity between the anatomical landmark template (\textit{i.e.} exemplar) and the target (\textit{i.e.} instance). As long as the ultrasound sequence becomes longer, the landmark position estimation is of more uncertainty and the accumulated tracking errors are larger \cite{wambersie1992prescribing}. In order to minimize the tracking errors, a system with the capability of learning deformable shapes is highly desirable.

\begin{figure*}[t]
    \centering
    \includegraphics[width=\textwidth]{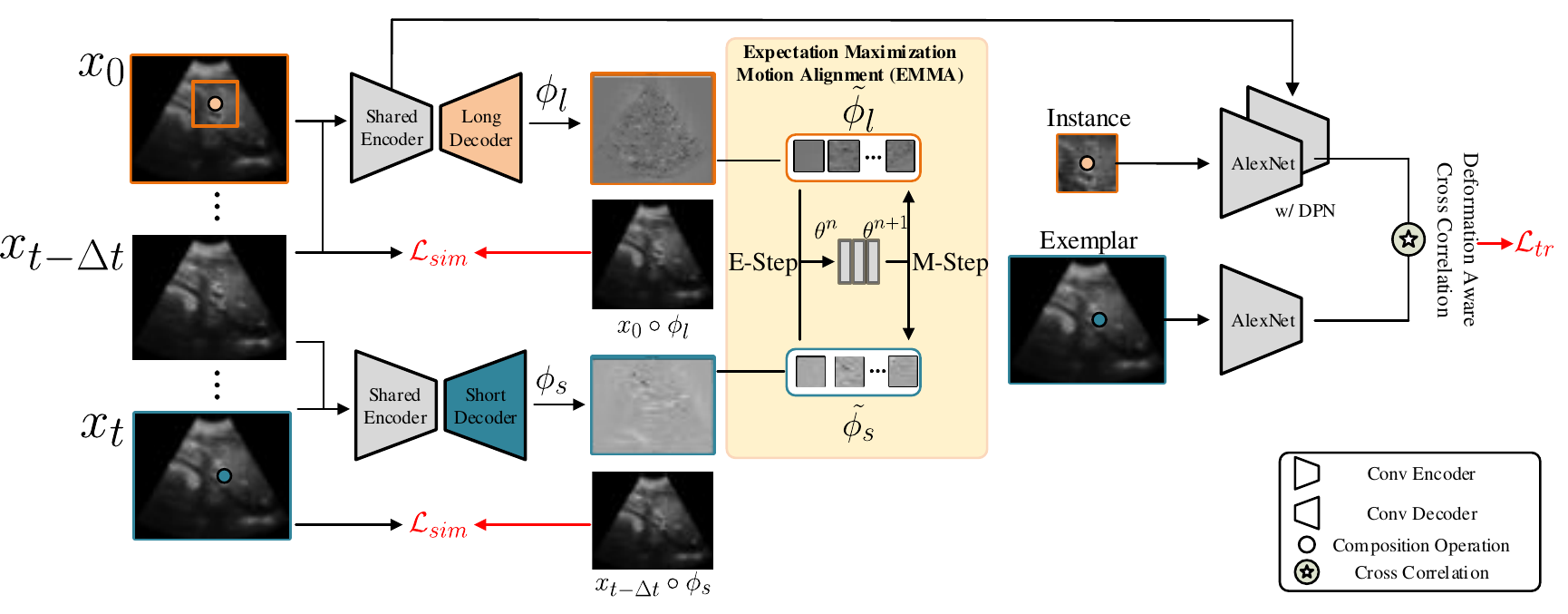}
    \caption{Pipeline of our proposed LSDM.}
\label{fig:pipeline}
\end{figure*}

In this paper, we propose a novel anatomical landmark tracking system with a learnable motion prior. Different from previous medical landmark tracking methods that only focused on the tracking task only, we design a multi-task tracking system using a novel diffeomorphism motion structure, namely long-short diffeomorphic motion (LSDM), to learn and adapt the non-rigid transformation between the target and the historic reference frames, generating a landmark motion prior in both long and short time intervals. LSDM can effectively explore the hybrid deformation and improve the downstream tracking task\textquotesingle s accuracy for searching the most plausible deformed landmark, eliminate noisy artifacts, which has not been studied before. We further design an expectation maximization motion alignment (EMMA) module to fully utilize the long interval temporal directional information and short interval spatial fine-grained information. In EMMA, the short deformation serves as a latent motion variable. The long deformation can be iteratively optimized by aligning to the motion representation of the same landmark. The optimized short deformation can also be recomposed and then form the long diffeomorphism with updated temporal details as a close-loop regularization to avoid over-fitting. Our contribution can be summarized as follows:

\begin{itemize}
    \item We propose a novel anatomical landmark tracking method by multi-tasking both tracking and diffeomorphism motion learning. Unlike previous methods that learn two tasks independently, we integrate the learned diffeomorphism as a motion prior for searching the best tracking candidate with significant deformation, thus to improve the tracking accuracy under challenging medical imaging situations such as landmark tracking in noisy background within ultrasound sequences.
    \item To effectively reduce the cumulative errors in long range modeling, we design a new representation structure containing diffeomorphism in both long and short time intervals (LSDM). We also propose an expectation maximization motion alignment (EMMA) module to iteratively update both long and short time deformation utilizing motion direction and similarities. Through EMMA, both long and short diffeomorphism can be updated by alternatively constructing and optimizing the lower bound of motion evidence, resulting in aligning deformations to the same directional and spatial representation, which further mitigates the motion biases brought by local anatomical ambiguity.
    \item LSDM can be trained within an end-to-end weakly-supervised fashion, which only requires few landmark annotations for tracking and zero annotation for long-short motion learning. We conducted extensive experiments on public and private ultrasound videos. Results show that our proposed system out-performs various fully-supervised tracking only methods with high interpretation and strong generalization across different conditions.
\end{itemize}

The rest of this paper is structured as follows. We first present a comprehensive review of related works (Sec.\ref{sec:related works}) in medical image tracking, motion estimation, and point out how LSDM differs from them. In Sec.\ref{sec:method}, we detail our proposed LSDM tracking network and EMMA module design with theoretical analysis. Experimental setups and analysis are presented in Sec.\ref{sec:experiments} and \ref{sec:results}, respectively. Finally, we conclude this paper in Sec.\ref{sec:Conclusion}.

\section{Related Works}
\label{sec:related works}
We first briefly review related works in medical landmark tracking and relevant technologies from nature scene object tracking in Sec.\ref{lr:part1}, which can be roughly divided into registration and feature mapping based methods. We also discuss some preliminary works within diffeomorphic motion modeling in Sec.\ref{lr:part2}. More discussion on related works can be found in Table S1, Supplementary A.

\subsection{Medical Landmark Tracking}
\label{lr:part1}

Accurate anatomical landmark tracking provides fundamental and crucial information for minimizing the spatial margins during radiation therapy, specifically using ultrasound images with substantial benefits such as fast acquisition, low cost and non-invasion \cite{antico2019ultrasound}. Previous attempts on medical landmark tracking can be roughly divided into two categories based on the matching formulation: \textbf{registration based tracking} and \textbf{feature mapping based tracking}. Early attempts tried to learn a registration framework for finding the spatial transformation between the template frame and the rest frames, thus the landmark position can be calculated using the affine matrix between the template and the target frame. Banerjee et al. \cite{banerjee2015combined} tracked the landmark by matching the global and local point set between two frames. Konig et al. \cite{konig2014non} proposed a registration method for calculating the registered landmark position with a normalized gradient field. These registration-based tracking frameworks focused on minimizing the global image registration errors, while the local anatomical errors cannot be corrected in time. Moreover, the manually designed features cannot fully represent the landmark deformation during the longitude evolution \cite{yang2011prediction}, \cite{jiao2021deep}.

Instead of implicitly calculating the landmark position using the affine matrix from global image registration, inspired by siamese networks \cite{bertinetto2016fully}, \cite{tao2016siamese}, recent works focus on automatically learning the high dimensional features and measuring the feature similarity between the exemplar and the follow-up instances directly. The exemplar and instances extracted from the same landmark should be similar in the feature space generated from the same deep network. Bharadwaj et al. \cite{bharadwaj2021upgraded} applied the basic siamese network tracking the ultrasound liver landmark. However, the network can be confused by other regions with similar visual representation and the fixed displacement prior may fail when the landmark moves out of field-of-view or is hard to be distinguished from the background. To obviate these difficulties, Wu et al. \cite{wu2022fusion} extended the siamese network by calibrating the intermediate features. Liu et al. \cite{liu2020cascaded} developed a cascaded network using two branches of a siamese network with different sizes of inputs. However, these methods suffer from learning the complex pair-wise relationship between the instance and the exemplar by iterating every frame pair, which is label intensive for medical image analysis tasks. Also, the hierarchical feature learning network lacks meaningful insights on minimizing the tracking error, particularly on quantifying the safety margin required by radiation therapy treatment planning to deliver planned doses \cite{van2004errors}, \cite{van2002inclusion}.

\subsection{Diffeomorphic Motion Modelling}
\label{lr:part2}
Diffeomorphism describes the non-linear transformation, measuring how topology, such as components and connected boundaries, is deformed between two time stamps \cite{beg2005computing}, \cite{ashburner2007fast}. During radiation therapy, diffeomorphic motion estimation adversely affects the planned irradiation of the target anatomy. Recent research works use deep networks to learn the stationary velocity field parameterized as a scaling and squaring layer to form the diffeomorphic deformation \cite{balakrishnan2019voxelmorph}, \cite{mok2020fast}. These research works greatly boosted the diffeomorphism registration performance, however, they all followed the pair-wise deformation learning, i.e. a fixed frame is chosen as template for the rest of the images to be pair-wise warped, which introduces additional biases such as interpolation asymmetry and the deformation computation complexity is proportional to the length of the image dataset. 

Different from previous attempts, in this work, we design a simple but effective diffeomorphism structure called long-short diffeomorphism motion (LSDM). The long diffeomorphism serves as the pair-wise motion learning from a sparse image set formed by the selected frames (i.e. keyframes for landmark tracking). The short diffeomorphism plays a role as group-wise motion to learn the deformation between the keyframe and their previous ${\Delta t}^{th}$ frame. This long-short structure can be viewed as a hybrid structure where the long diffeomorphism from a sparse set can reduce the deformation complexity for a long video sequence and the short diffeomorphism from the paired images is invertible and able to preserve anatomical topology. Similar to the use of expectation maximization (EM) for learning the optimized parameters of mixture models, we propose an expectation maximization motion alignment (EMMA) module to iteratively update and find the best long-short deformation. Thus the drifting error caused by image artifacts can be effectively reduced by considering deformation in both local and global contexts along the temporal dimension. The optimized long-short diffeomorphism motion can be served as a complete prior for downstream tasks, where we show the LSDM can provide meaningful deformation information for feature extraction networks used in a relatively challenging task such as ultrasound landmark tracking.

\section{Method}
\label{sec:method}
\subsection{Problem Formulation}
We focus on anatomical landmark tracking within ultrasound image sequences. The goal is to generate precise location estimation for subsequent frames given the starting position of the landmark, which requires a stable single object-level description with high efficacy and low complexity. Thus, we formulate the ultrasound anatomical landmark tracking problem following the single object tracking definition based on similarity learning \cite{bertinetto2016fully, tao2016siamese}. The similarity-based tracking aims to estimate the location $s$ of landmark (exemplar) $x_{0}$ in the $t$-th tracking frame (instance) $x^{s}_{t}$ using an optimized siamese network $\Phi$ parameterized by $\theta$, resulting in minimizing the following tracking loss function:

\begin{equation}
\label{eqn:definition}
    \mathcal{L}_{Tr}\left(\theta\right) = \sum_{t} \| \Phi(x_{0}) \star \Phi(x^{s}_{t}) - y^{s}_{t} \|^{2}
\end{equation}

where $y^{s}_{t}$ is the associated true Gaussian confidence map at location $s$ in the $t$-th frame. $\star$ represents the correlation operation. As we summarized in Sec.\ref{sec:related works}, previous attempts for finding the feature similarity are heavily affected by internal and external noise. Moreover, previous tracking models cannot construct a morphology search space to measure a precious deformation distance between the exemplar and the instance, missing an important prior for accurate anatomical landmark tracking. Following multi-task setting \cite{zhang2021survey}, we propose a novel framework for learning a unified filter $\Phi_{U}$ for both tracking landmark $x_{0}^{s}$ at locations $\boldsymbol{s} = \left\{ s_0, s_1, ..., s_t \right\}$ and generating diffeomorphic deformation $\boldsymbol{m} = \left\{m_0, m_1, ..., m_t\right\}$ to update the search template with most plausible diffeomorphic transformation, a given dataset $\boldsymbol{x}$ with $t$ frames $\boldsymbol{x}=\left\{x_0, x_1,..., x_t\right\}$:

\begin{equation}\label{eqn:objective}
\begin{split}
    \Phi^{\star}_{U} &= \argmaxA_{\Phi_{U}}p \left(\boldsymbol{s}, \boldsymbol{m} \middle | \boldsymbol{x}, \Phi_{U}\right)\\
        &= \argmaxA_{\Phi_{U}}p\left(\boldsymbol{s} \middle | \boldsymbol{m}, \boldsymbol{x}, \Phi_{U}\right)p\left(\boldsymbol{m} \middle |\boldsymbol{x}, \Phi_{U}\right)
\end{split}
\end{equation}

Following Eqn.(\ref{eqn:objective}), we formulate the landmark tracking as an optimization problem with diffeomorphic transform estimation as learnable equality constraints. We expect to find the optimal unified filter $\Phi^{\star}_{U}$ to minimize the landmark location tracking error $\mathcal{L}_{Tr}$ and the diffeomorphic transform error $\mathcal{L}_{M}$ simultaneously, given a dataset $\boldsymbol{x}$ which contains $t$ frames with a landmark (instance) $x_{F}^{s}$ a fixed frame $x_{F}$ (here we refer the first frame $x_{0}$ as $x_{F}$) as transformation template:

\begin{equation} \label{eqn:optimization}
\begin{split}
\min \mathcal{L}_{Tr}& =  \sum_{t} \| \Phi_{U}\left(x_{0}^{s}\right) \star \Phi_{U}\left(x_t\right) - y_{t} \|^{2}\\
\text{s.t}~\mathcal{L}_{M} &= 0\\
\text{where}~\mathcal{L}_{M} &= \sum_{t}\| \Phi_{U}\left(x_{0}, x_{t}\right)-x_{t} \|^{2}
\end{split}
\end{equation}

\subsection{Long-Short Diffeomorphic Motion Structure}
A na\"ive solution for Eqn.(\ref{eqn:optimization}) is to learn a single diffeomorphic motion that can estimate an accurate transformation between template frame $x_F$ and target frame $x_t$. This diffeomorphic prior can benefit the jointly trained tracker, where the tracker cannot only learn the feature similarity, but also search the most plausible location of instance $x_{0}^{s}$ in the target frame $x_t$ in the motion space generated from the diffeomorphic transformation $\Phi_{U}\left(x_0, x_t\right)$. However, one main drawback of this na\"ive setting is the cumulative error within a long tracking sequence cannot be effectively corrected, i.e. multiple adversarial response peaks with noisy motion prediction cannot be erased, resulting in a high variance of tracking trajectory along the temporal domain, which is not ideal for minimizing the anatomical landmark location margin during radiation therapy. Last but not least, minor motion changes cannot be detected as the unified filter tends to learn a complete motion for the whole sequence while the motion shift between a few frames could be ignored.

Inspired by classical findings in video temporal analysis \cite{adelson1985spatiotemporal, feichtenhofer2019slowfast}, we propose a novel diffeomorphic representation structure called long-short diffeomorphism motion (LSDM) to solve the issues listed above. LSDM contains estimation of diffeomorphic motion $\Phi_{U} = \left\{\boldsymbol{\phi}_{l}, \boldsymbol{\phi}_{s}, \boldsymbol{\phi}_{Tr}\right\}$ in frame pairs with both long time interval $t$ and a stochastic short time interval $\Delta t$ ($\Delta t \ll t$). We follow the standard definition of diffeomorphism by assuming the latent variable $z$ of both long and short diffeomorphism is a multivariate Gaussian distribution with zero mean and covariance $\sigma$:

\begin{equation}
    p\left( z \right) \sim \mathcal{N}(0, \sigma^{2})
\end{equation}

where $z$ is a stationary velocity field (SVF) generated by deformation field $\boldsymbol{\phi}$ within time interval $t$ $\in$ $[0, 1]$:

\begin{equation}
\label{eqn:velocity}
    \frac{d \boldsymbol{\phi}^{t}}{dt} = \boldsymbol{v}(\boldsymbol{\phi}^{(t)}) =  \boldsymbol{v} \circ \boldsymbol{\phi}^{(t)}
\end{equation}

where $\circ$ is the composition operator. Different from traditional diffeomorphism learning given a single fixed template $x_{F}$, the template of long diffeomorphism is fixed as the first frame of the frame sequence $x_{F} = x_0$ and the template of short diffeomorphism is sampled $x_{F} = x_{t-\Delta t}$ based on the short time interval $\Delta t$. Thus, we can obtain a noisy observation of the warped images set: $\left\{x_0 \circ \boldsymbol{\phi}_{l}, x_{t-\Delta t} \circ \boldsymbol{\phi}_{s} \right\}$ from a Gaussian mixture model:

\begin{equation}
\begin{split}
    p(y_{t-\Delta t}, y_{t} | z;x_0, x_{t- \Delta t}) & \propto \mathcal{N}(y_{t-\Delta t};x_0 \circ \boldsymbol{\phi}_{l}, \sigma^{2} \mathbb{I}) + \\ &\mathcal{N}(y_{t};x_{t - \Delta t} \circ \boldsymbol{\phi}_{s}, \sigma^{2} \mathbb{I})
\end{split}
\end{equation}

where $y_{t-\Delta t}$ and $y_{t}$ are the observations of the warped images given $x_0$ and $x_{t - \Delta t}$ respectively. By introducing the hierarchical structure of diffeomorphism in both long and short temporal domains, we are able to learn a reliable diffeomorphism combination. The long time diffeomorphism can benefit the downstream tracker to minimize the spatial search space for finding the exemplar $x^{s}_{0}$. The short diffeomorphism allows us to effectively learn the preservation of short time deformation changes based on adaptive updated template $x_{t - \Delta t}$ and can benefit the tracker with the updated motion trends from frame $x_{t - \Delta t}$ to frame $x_{t}$.

\begin{figure}
    \centering
    \includegraphics[width=0.48\textwidth]{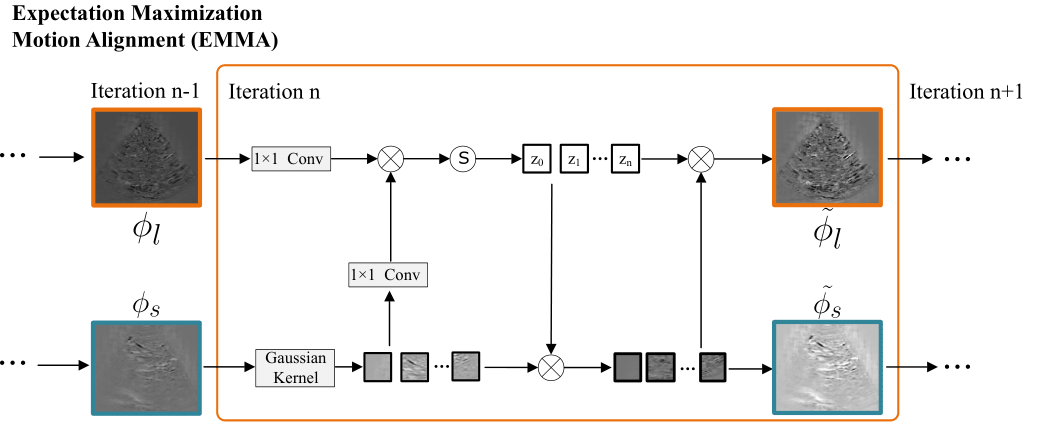}
    \caption{Illustration of the proposed Expectation Maximization Motion Alignment (EMMA) module. Cross means element-wise multiply. ``S'' means sigmoid layer. EMMA first kernelizes the input deformation, where long (orange) and short (blue) range deformation with iterative updates are shown in Alg. \ref{Alg:convmf}. The output of EMMA is the aligned long and short range deformation, which benefits the downstream trackers to search the tracking instance with significant deformation.}
\label{fig:emma}
\end{figure}

\subsection{Expectation Maximization Motion Alignment}
The long and short diffeomorphism deformation focuses on learning different parts of the deformation due to the choice of the template being different. The template of the long deformation is fixed (as $x_0$), causing the long deformation to focus on the generalized motion learning with a long interval motion trajectory. The template of short diffeomorphism is adapted, which leads the short deformation to seeking an average mapping based on the finding of the similarity between $x_{t - \Delta t}$ and $x_t$. 

Considering either long or short diffeomorphic motion only may lead to biased deformation curve estimation from a sparse partially prior (shown in Fig. \ref{fig:longshort}). In practice, finding corresponding image pixels over long diffeomorphic motion is not trivial at all. First of all, it is extremely difficult to identify optimal correspondences in a noisy image space over a long interval. Second, based on Bayesian inference, physics driven prediction models likely suggest several similar candidates in the following image frames, which may further confuse the downstream tracking algorithm. Finally, instead of rendering corresponding counterparts in the next image frame over a long interval, we rather target at a direct approximation of image pixels over continuous short diffeomorphic motion intervals. One of the possible solutions is to use the expectation-maximization algorithm. In this paper, we propose an Expectation Maximization Motion Alignment (EMMA) algorithm to align both long and short deformation. Specifically, at the $n^{th}$ iteration, the short motion is an extension of long motion in both spatial-temporal domains, i.e., the direction and distance distribution of short motion between $x_{t - \Delta t}$ and $x_t$ follows the moving inertia of long motion between $x_{0}$ and $x_{t}$. Our proposed EMMA follows this physical model by viewing the short motion as a latent motion vector that generates the corresponding long motion following a conditional posterior distribution $p(\phi_{s} | \phi_{l}, \theta)$, parameterized by $\theta$:

\begin{equation}
\label{eqn:theta}
    \theta^{\star} = \argmaxA_{\theta} \log p(\phi_l | \phi_s, \theta)
\end{equation}

where $\phi_s$ is the short motion used as latent variable and together $(\phi_l, \phi_s)$ is called complete motion. Our goal is that by using the iterative optimization method EMMA, we can find the optimal parameter $\theta^{\ast}$ for generating the final long motion representation, where $\theta^{\ast}$ for Eqn. \ref{eqn:theta} can be solved by the following theorem.

\begin{theorem}
\label{the:emma convergence}
The expectation maximization motion alignment can be converged with an optimum $\theta^{\ast}$ within $N$ times iteration where:
\begin{equation}
    \log p(\phi_l | \phi_s, \theta^{\ast}) \geq \log p (\phi_l | \phi_s, \theta^{n}), n = 1, …, N, N \rightarrow \infty
\end{equation}
\end{theorem}

We provide the proof of Theorem \ref{the:emma convergence} in Supplementary B, which leads to the E-step and M-step of EMMA. Specifically, EMMA first starts with random initialized parameter $\theta$ and constructs the expectation expression at iteration $n$:
\begin{equation}
\label{eqn:estep}
\begin{split}
    \text{E-step:}&\\
    &p(\phi_s|\phi_l, \theta^{n}) \rightarrow E_{\phi_s|\phi_l, \theta^{t}}\left [\ \log p(\phi_l, \phi_s|\theta) \right ]\
\end{split}
\end{equation}
and update the parameter for maximizing the constructed expectation:
\begin{equation}
\label{eqn:mstep}
\begin{split}
    \text{M-step:}&\\
    &\theta^{n+1} = \argmaxA_{\theta} E_{\phi_s|\phi_l, \theta^{n}} \left [\ \log p(\phi_s, \phi_l|\theta) \right ]\
\end{split}
\end{equation}

\begin{algorithm}[t]
\caption{Expectation Maximization based Long Short Deformation Alignment.}
\begin{algorithmic}[1]
\renewcommand{\algorithmicrequire}{\textbf{Input:}}
\renewcommand{\algorithmicensure}{\textbf{Output:}}
\REQUIRE Generated long deformation $\phi_l$ and short deformation $\phi_s$.
\ENSURE Aligned deformation $\phi_l^N$ and $\phi_s^N$ after N iterations.
\STATE Kernelize $\phi_s$ to $\tilde{\phi_s}$
\WHILE {iteration stage $<$ N}
 \STATE $\phi_s^N = \phi_l * \tilde{\phi_s}$;
 \STATE $z = \text{Softmax}(\phi_s^N)$
 \STATE $\phi_l^N = z * \tilde{\phi_s}$
\ENDWHILE
\STATE $\phi_s^N = \phi_l^N * z$
\STATE $\phi_s^N = \phi_s^N + \phi_s$
\STATE $\phi_l^N = \phi_l^N + \phi_l$
\RETURN Aligned long deformation $\phi_l^N$ as $\phi_l$, short deformation $\phi_s^N$ as $\phi_s$.
\end{algorithmic}
\label{Alg:convmf}
\end{algorithm}

\subsection{Plug-and-Play Differentiable EMMA}
Fomulated from E-Step (Eqn.(\ref{eqn:estep})) and M-Step (Eqn.(\ref{eqn:mstep})), EMMA first kernelized short diffeomorphism $\phi_{s}$ into $k$ seed features $\tilde{\phi_{s}} = \{ \phi_{s}^{1},..., \phi_{s}^{k}\}$. We model the conditional distribution of $\phi_{l}$ and $\tilde{\phi_{s}}$ as a Gaussian Mixture Model as the short diffeomorphism $\phi_{s}$ is an extent of $\phi_{l}$ in the temporal domain, parameterized by latent variable $Z = \{ z_{\phi}^{1}, ..., z_{\phi}^{k} \}$. 

\begin{equation}
\label{eqn:elbo}
    p(\phi_{s} | \phi_{l}) = \sum_{k=1}^{K}z_{\phi}^{k}N(\phi_{s} | \phi_{l} ; \mu^{k}, \sigma^{k})
\end{equation}

The log-likelihood of $\phi_{l}$ and $\phi_{s}$ given a fixed parameter $\theta$ can be written as:

\begin{equation}
\label{eqn:elbo}
\begin{split}
    \log p(\phi_{s}, \phi_{l};\theta) &= \log \sum_{z_{\phi}}p(\phi_{l}, \phi_{s}, z_{\phi};\theta)\\
    &= \log \sum_{z_{\phi}}Q(z_{\phi})\frac{p(\phi_{l}, \phi_{s}, z_{\phi};\theta)}{Q(z_{\phi})}\\
    &\geq \sum_{z_{\phi}}Q(z_{\phi}) \log \left \{ \frac{p(\phi_{l}, \phi_{s}, z_{\phi};\theta)}{Q(z_{\phi})} \right \}
\end{split}
\end{equation}

where

\begin{equation}
\begin{split}
Q(z_{\phi}) &= \frac{p(\phi_{l}, \phi_{s}, z_{\phi};\theta)}{\sum_{z_{\phi}}p(\phi_{l}, \phi_{s}, z_{\phi};\theta)}\\
&= \frac{p(\phi_{l}, \phi_{s}, z_{\phi};\theta)}{p(\phi_{l}, \phi_{s};\theta)}\\
&=p(z_{\phi} | \phi_{l}, \phi_{s};\theta)
\end{split}
\end{equation}

Based on the Expectation-Maximization algorithm, the E-step of EMMA is to construct a motion evidence lower bound (ELBO) for $log~p(\phi_{s}, \phi_{l};\theta)$ in Eqn.(\ref{eqn:elbo}) and the M-step aims to maximizing the ELBO w.r.t $\theta$ while fixing $Q(z_{\phi})$, which means to maximize the likelihood of $\phi_{s}$ for the next time step given long deformation $\phi_{l}$ generated from the history, specifically:

\begin{equation}
\begin{split}
    Q(z_{\phi}^{i}) &= \sum_{i=1}^{k}p(z_{\phi}^{i}|\phi_l, \phi_s, \theta)\\
    z_{\phi}^{k} &= \frac{ \Ker (\phi_{l}, \tilde{\phi_{s}^{k})}}{\sum_{j=1}^{k} \Ker (\phi_{l}, \tilde{\phi_{s}^{j}})}\\ 
\end{split}
\end{equation}
where $\Ker$ is the kernel mapping function for both $\phi_l$ and $\phi_s$. We apply the simple inner product function as the choice of $\Ker$ for the convenience of implementation. Following previous works in \cite{li2019expectation}, \cite{wang2018non}, we implement the E-step as a softmax layer for the inner product of $\phi_l$ and $\tilde{\phi_s}$ to formulate the probabilistic distribution of $Z$:
\begin{equation}
    Z = \text{Softmax}(\phi_l (\tilde{\phi_s}) ^ {\top})
\end{equation}

After we have fixed the latent variable $Z$, the M-step of EMMA is to maximize the probability of the appearance of kernelized short diffeomorphism feature given a previous long diffeomorphism:
\begin{equation}
    \tilde{\phi_s} = \frac{z_{\phi}^{i}\phi_{l}}{\sum_{j=1}^{k}z_{\phi}^{j}}
\end{equation}
After the iteration of $N$ times, $\tilde{\phi_{s}}$ can be updated to an optimal representation. Following the definition of diffeomorphic learning in Eqn.(\ref{eqn:velocity}), the long deformation $\phi_{l}$ can be re-composed using the optimized $\tilde{\phi_{s}}$, as the optimized short diffeomorphism aligns with the same snapshot within a single time slot:
\begin{equation}
    \phi_{l}^{N} = z_{\phi} \tilde{\phi_{s}^{N}}
\end{equation}
The re-composition of $\phi_l$ can be viewed as a regularization strategy. After having finished the re-composition, EMMA achieves a close loop updating $\phi_l$ and $\phi_s$. The algorithm of executing EMMA for $N$ iterations can be summarized as Alg.\ref{Alg:convmf}. Since the common frame $x_{t - \Delta t}$ is involved during EMMA, we believe that the updated $\phi_s^N$ and $\phi_l^N$ are aligned together, following the definition of feature alignment in similar research fields such as domain-adaptation \cite{shen2020multi} and object re-identification \cite{zhang2017alignedreid}.

\begin{figure}
    \centering
    \includegraphics[width=0.48\textwidth]{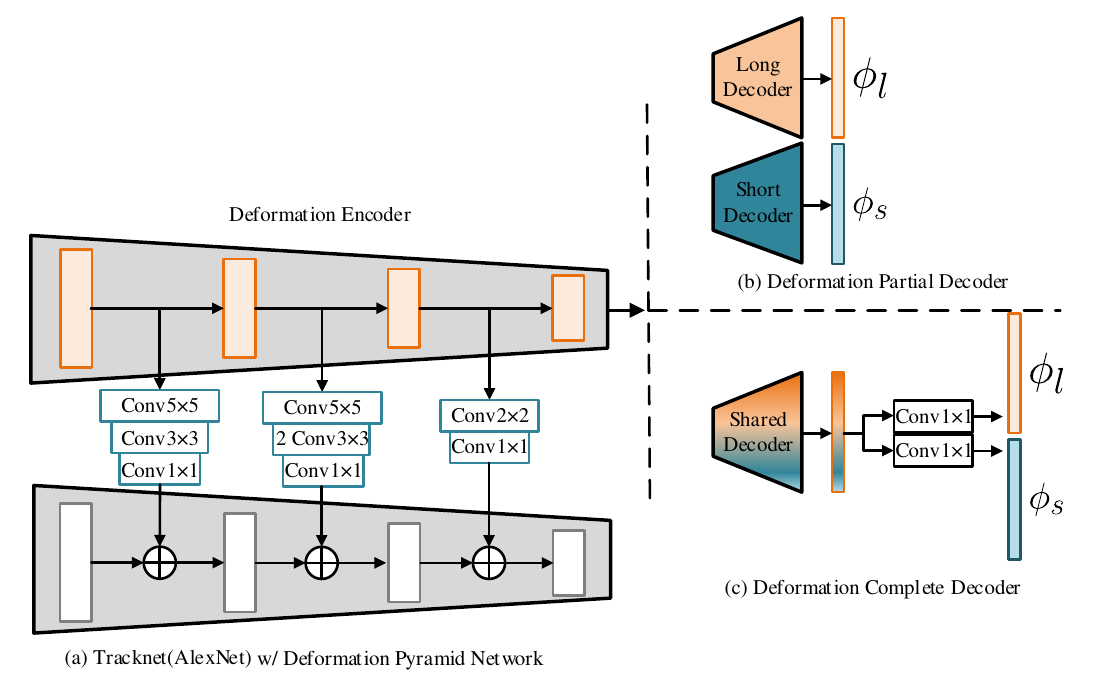}
    \caption{Illustration of the proposed Deformation Pyramid Network (DPN). Left (a): the deformation pyramid network transferring the motion feature (orange) as a prior to the tracking network (gray). Right (b): Partial motion decoder design, where long and short range deformation is modeled by two separate decoders. Right (c): Complete motion decoder design, where long and short range deformation is modeled by the same decoder and only decoupled at the last layer.}
\label{fig:dpn}
\end{figure}

\subsection{Deformation Pyramid Network}
Inspired by Feature Pyramid Network (FPN) \cite{lin2017feature}, we take the feature from long-short diffeomorphism encoding as a prior to learn a plausible deformation for tracking an instance for the exemplar in subsequent frames. Specifically, a traditional siamese network uses a shared encoder to extract features for both the exemplar and the instance. The shared weights setting encourages learning a high similarity, while ignoring the exemplar\textquotesingle s deformation along the spatial-temporal dimension, which is significant for medical image tracking. Here, we take the intermediate features from the deformation network as a prior for the exemplar, fused by the proposed Deformation Pyramid Network (DPN). Similar to FPN, DPN takes the deformation feature hierarchy with semantics from low to high levels, to update the exemplar features at the corresponded levels. 

As shown in Fig. \ref{fig:dpn}, the DPN follows a top-down construction, where the feature maps with larger size are down-sampled by large sized kernels and smaller size feature maps are down-sampled by small sized kernels. Note that there are two ways to generate long and short deformation based on how the long and short deformation share their networks\textquotesingle ~weight. The first one is called complete network, where the long and short deformation share the encoder and the decoder of the entire deformation learning network. In the complete network, the long and short deformation can only be separated at the last layer of the decoder. Another set up is called partial network, where the long and short deformation shares the encoder of the deformation network. In the partial network, the two independent decoders are used to generate long and short deformation respectively. We test the effects of DPN with inputs from different deformation network structures (complete vs.\ partial), detailed in Sec.\ref{sec:ablation}. The structural test does not show any preference of complete vs. partial. For the trade-off between computational cost and performance, we choose the complete network design.

\section{Experiments}
\label{sec:experiments}
We demonstrate the effectiveness of boosting ultrasound anatomical landmark tracking performance with multi-tasking diffeomorphism prior by conducting extensive experiments on two datasets: 2D ultrasound video dataset from public Challenge on Liver Ultrasound Tracking (CLUST2D) and private collected 2D kidney ultrasound video dataset from the Affiliated Hospital of Nantong University with normal and contrast ultrasound modality (AHNTU2D).

\begin{figure*}[thp]
    \centering
    \includegraphics[width=\textwidth]{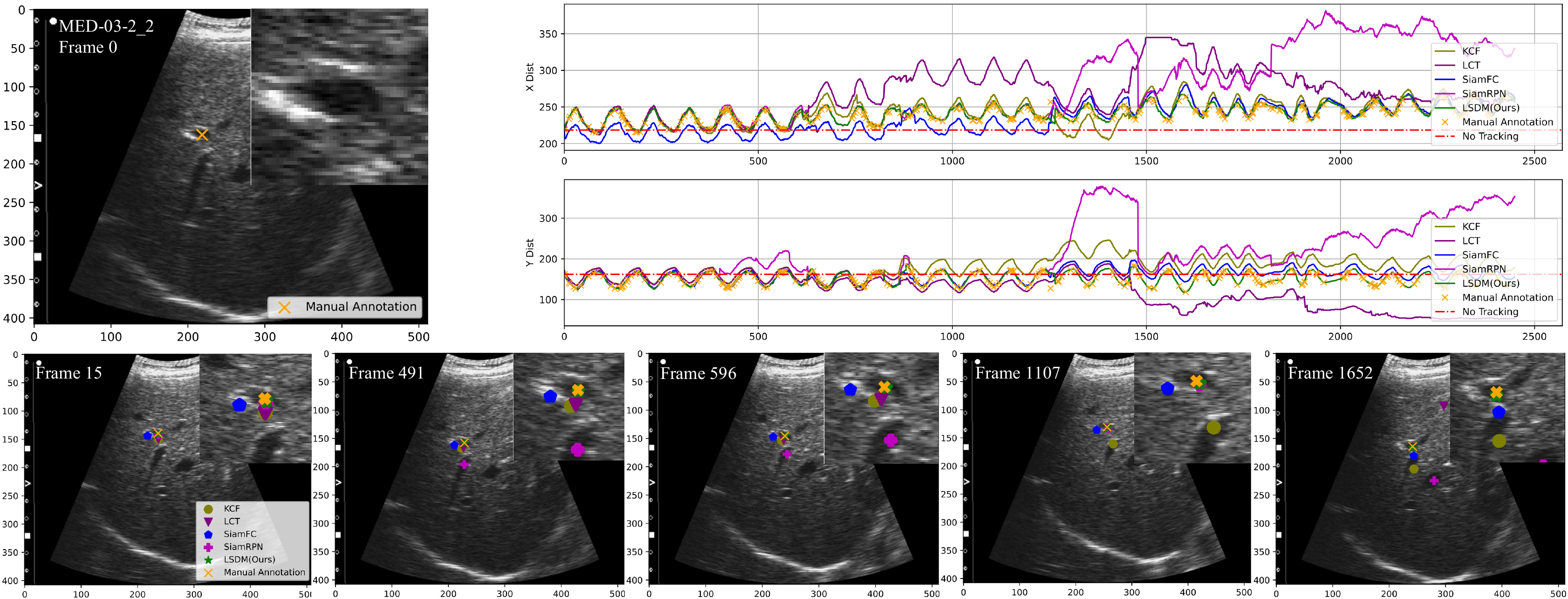}
    \caption{Qualitative comparison of different baseline models and the proposed LSDM by tracking validation on the CLUST2D training set. We select a representative dataset entry to demonstrate LSDM superior tracking performance. We show the start frame at frame 0 (top-left), landmark center coordinate tracklet comparison (top-right) and selected frames with landmark tracking results (bottom). We can observe a stable and accurate tracking performance of LSDM, while other baselines fail at accumulating tracking error through the whole sequence. Supplementary V summarizes other tracking results and their visualization. Best viewed in color.}
\label{fig:trackresult}
\end{figure*}

\begin{table*}[h]
\caption{Quantitative comparisons between LSDM and other representative tracking techniques on CLUST2D training set with respect to mean, standard deviation, 95th, min and max tracking error. In `Feature' column, `M' means `manual extraction' and `A' means `automatic extraction'.}
\small
\centering\centering
\begin{tabular}{|l|c|c|c|c|c|c|c|c|}
\hline
\multicolumn{1}{|c|}{Method} & Feature & Deformation & Long Short Memory & Mean & Std  & 95th  & Min  & Max   \\ \hline
KCF \cite{henriques2014high}                          & M       & \ding{55}   & \ding{55}       & 9.35 & 3.67 & 19.21 & 0.09 & 67.92 \\ \hline
LCT \cite{ma2018adaptive}                         & M    & \ding{55}   & \ding{51}       & 5.12 & 2.54 & 8.3   & 0.02 & 48.82 \\ \hline
SiamFC \cite{bertinetto2016fully}                       & A       & \ding{55}   & \ding{55}      & 4.39 & 2.29 & 5.33  & 0.01 & 33.76 \\ \hline
SiamRPN \cite{li2018high}                      & A       & \ding{55}   &  \ding{55}      & 2.42 & 1.86 & 4.26  & 0.01 & 18.37 \\ \hline
LSDM(Ours)                   & A       & \ding{51}   &  \ding{51}      & 0.81 & 0.98 & 1.73  & 0.01 & 14.92 \\ \hline
\end{tabular}
\label{Table:Training set result}
\end{table*}

\subsection{Dataset}
CLUST2D is composed of 43 patients with different times of acquisition, a total of 63 ultrasound videos collected from 4 different groups, where 24 videos are used for training and 39 videos are used for testing. In the training set, the length of the 2D video sequence varies from 1075 to 5247 frames, with each frame resolution varying from 393 $\times$ 457 to 524 $\times$ 591. Landmarks to be tracked for each video subject in the training set varies from 1 to 5. In the test set, the length of the 2D video sequence varies from 895 to 15640, with each frame resolution varying from 262 $\times$ 313 to 524 $\times$ 591. The number of landmarks to be tracked for each video in the test set varies from 1 to 4. Training labels (landmark coordinate x, y) are provided by reliable observers. Roughly $10\%$ of the total sequence is annotated with random time intervals between two entries. Test set annotation is only provided for the first frame. Generated tracking results on the test set will be submitted and evaluated at the official server. We suggest the reader check \cite{de2018evaluation} and official challenge \text{website}\footnote{https://clust.ethz.ch/} for more detailed information.

AHNTU2D is composed of 12 acute kidney injury patients with different times of acquisition, totally 14 ultrasound videos focused on both kidney collected from the Affiliated Hospital of Nantong University, China. For each video sequence, AHNTU2D contains two modalities, i.e. normal ultrasound (\text{AHNTU2D-N}) and contrast ultrasound (\text{AHNTU2D-C}). For both \text{AHNTU2D-N} and \text{AHNTU2D-C}, we randomly allocate 6 videos for training, 4 videos for validating and 4 videos for testing. The spatial resolution is 550 $\times$ 900 and the temporal length varies from 280 to 3966. Landmarks to be tracked for each video sequence varies from 1 to 3. All videos are manually labelled by one experienced physician from the Affiliated Hospital of Nantong University. The data collection followed the ethic procedure of the Affiliated Hospital of Nantong University. All video sequences have been desensitized following the standard procedure. 

\subsection{Evaluation Metric}
Euclidean distance is used for evaluating the tracking error (TE) on the $i^{th}$ landmark between the predicted center coordinate $P_{t}^{i}$ and the ground truth coordinate $GT_{t}^{i}$ at $t^{th}$ frame.
\begin{equation}
	TE_t^{i} = ||P_t^i – GT_t^i||
\end{equation}
TE is evaluated and summarized with mean, standard deviation and 95th percentile from the official evaluation server. For system evaluation with motion magnitude, no tracking error (NoTE) is included as standard calculation:

\begin{equation}
	NoTE_t^i= ||P_0^i – GT_t^i||
\end{equation}

\subsection{Baseline Methods}
We first the effectiveness of the deformation prior, EMMA and DPN with ablation studies. In-House testing is conducted as well, where the validation sequence provided by a specific hospital is unseen during the training. We also test the stabilization performance of LSDM on a randomly split training set.

We also present the systematic tracking performance against several state-of-the-art ultrasound landmark tracking methods, including 2D tracking methods \cite{liu2020cascaded}, \cite{shepard2017block}, \cite{williamson2018ultrasound}, \cite{wu2022fusion}, \cite{shen2019discriminative} with leading performance on CLUST2D test set. As we built our tracking backbone upon the similarity learning, we also re-implement the state-of-the-art tracking method  SiamFC \cite{bertinetto2016fully} and SiamRPN \cite{li2018high} for further comparison.

\section{Results}
\label{sec:results}
In this section, we present our experimental design and result analysis. We first test the effectiveness of different components within LSDM in Ablation Study (Sec.\ref{sec:ablation}), leading to the combination of LSDM. Then we evaluate the generalization and robustness against different data providers with different ultrasound scanners in In-House Test (Sec.\ref{sec:inhouse}), which has never been carried out in previous works. We also compare our method against several representative tracking methods on the randomly split training set with ground truth (Sec.\ref{sec:trainingset}). Finally, we report the test set result on CLUST2D, AHNTU2D-N and AHNTU2D-C against other state-of-art methods, human expert observation and no-tracking results (Sec.\ref{sec:testset}). We also add failure tracking cases and fully analyze the reasons of mistracking (Sec.\ref{sec:FailureCase}).

\begin{table*}[h]
\label{Table:Groupwise SOTA}
\caption{Quantitative group-wise comparisons between LSDM and the other state-of-the-art methods on CLUST2D test set with respect to Mean, Standard deviation and 95th Tracking Error. The best result is shown in bold and the runner-up result is underlined.}

\begin{adjustbox}{width=1\textwidth}

\begin{tabular}{|l|ccc|ccc|ccc|ccc|ccc|}
\hline
\multicolumn{1}{|c|}{\multirow{2}{*}{CLUST2D}} & \multicolumn{3}{c|}{CIL}                                                               & \multicolumn{3}{c|}{ETH}                                                               & \multicolumn{3}{c|}{ICR}                                                               & \multicolumn{3}{c|}{MED1}                                                              & \multicolumn{3}{c|}{MED2}                                                              \\ \cline{2-16} 
\multicolumn{1}{|c|}{}                  & \multicolumn{1}{c|}{Mean}          & \multicolumn{1}{c|}{Std}           & TE95th       & \multicolumn{1}{c|}{Mean}         & \multicolumn{1}{c|}{Std}           & TE95th        & \multicolumn{1}{c|}{Mean}          & \multicolumn{1}{c|}{Std}          & TE95th        & \multicolumn{1}{c|}{Mean}          & \multicolumn{1}{c|}{Std}          & TE95th        & \multicolumn{1}{c|}{Mean}         & \multicolumn{1}{c|}{Std}           & TE95th        \\ \hline
Liu et al. \cite{liu2020cascaded}                              & \multicolumn{1}{c|}{1.19}          & \multicolumn{1}{c|}{1.16}          & 4.16         & \multicolumn{1}{c|}{\underline{ 0.59}}   & \multicolumn{1}{c|}{\underline{ 0.57}}    & \underline{ 1.24}    & \multicolumn{1}{c|}{\textbf{0.77}} & \multicolumn{1}{c|}{0.78}         & 2.70           & \multicolumn{1}{c|}{0.78}          & \multicolumn{1}{c|}{\textbf{0.60}} & \textbf{1.81} & \multicolumn{1}{c|}{\textbf{0.80}} & \multicolumn{1}{c|}{0.90}           & \textbf{1.73} \\ \hline
Williamson et   al. \cite{williamson2018ultrasound}                     & \multicolumn{1}{c|}{\textbf{1.01}} & \multicolumn{1}{c|}{\underline{ 0.84}}    & \underline{ 2.81}   & \multicolumn{1}{c|}{\textbf{0.50}} & \multicolumn{1}{c|}{\textbf{0.45}} & \textbf{1.13} & \multicolumn{1}{c|}{1.06}          & \multicolumn{1}{c|}{2.02}         & \underline{ 2.45}    & \multicolumn{1}{c|}{\underline{ 1.04}}    & \multicolumn{1}{c|}{1.14}         & 2.75          & \multicolumn{1}{c|}{0.99}         & \multicolumn{1}{c|}{0.78}          & 2.75          \\ \hline
Wu et al. \cite{wu2022fusion}                               & \multicolumn{1}{c|}{1.41}          & \multicolumn{1}{c|}{1.29}          & 4.20          & \multicolumn{1}{c|}{0.62}         & \multicolumn{1}{c|}{0.79}          & 1.46          & \multicolumn{1}{c|}{0.87}          & \multicolumn{1}{c|}{1.01}         & 3.30           & \multicolumn{1}{c|}{\textbf{1.02}} & \multicolumn{1}{c|}{1.58}         & 2.78          & \multicolumn{1}{c|}{1.10}          & \multicolumn{1}{c|}{1.64}          & 3.06          \\ \hline
Bharadwaj et   al. \cite{bharadwaj2021upgraded}                     & \multicolumn{1}{c|}{1.17}          & \multicolumn{1}{c|}{0.89}          & 2.95         & \multicolumn{1}{c|}{1.65}         & \multicolumn{1}{c|}{4.48}          & 2.65          & \multicolumn{1}{c|}{1.29}          & \multicolumn{1}{c|}{1.83}         & 5.16          & \multicolumn{1}{c|}{1.74}          & \multicolumn{1}{c|}{2.93}         & 5.80           & \multicolumn{1}{c|}{1.57}         & \multicolumn{1}{c|}{1.93}          & 6.72          \\ \hline
Shen et al. \cite{shen2019discriminative}                             & \multicolumn{1}{c|}{1.25}          & \multicolumn{1}{c|}{1.15}          & 4.03         & \multicolumn{1}{c|}{0.98}         & \multicolumn{1}{c|}{0.60}           & 2.16          & \multicolumn{1}{c|}{1.02}          & \multicolumn{1}{c|}{\underline{ 0.73}}   & 2.61          & \multicolumn{1}{c|}{1.54}          & \multicolumn{1}{c|}{1.49}         & 4.76          & \multicolumn{1}{c|}{1.04}         & \multicolumn{1}{c|}{\underline{ 0.67}}    & 2.18          \\ \hline
LSDM (Ours)                             & \multicolumn{1}{c|}{\underline{ 1.10}}     & \multicolumn{1}{c|}{\textbf{0.79}} & \textbf{2.70} & \multicolumn{1}{c|}{1.03}         & \multicolumn{1}{c|}{1.38}          & 2.19          & \multicolumn{1}{c|}{\underline{ 0.86}}    & \multicolumn{1}{c|}{\textbf{0.70}} & \textbf{1.86} & \multicolumn{1}{c|}{1.08}          & \multicolumn{1}{c|}{\underline{ 0.93}}   & \underline{ 2.55}    & \multicolumn{1}{c|}{\underline{ 0.89}}   & \multicolumn{1}{c|}{\textbf{0.59}} & \underline{ 1.99}    \\ \hline
\end{tabular}
\label{table:grouptest}
\end{adjustbox}

\end{table*}

\subsection{Ablation Study}
\label{sec:ablation}
We evaluate the impacts of different components within LSDM on the CLUST2D training set. The result is shown in Supplementary C. We first evaluate the deformation quality brought by different designs of complete and partial diffeomorphism learning. We then validate that EMMA can upgrade the lower bound of long deformation by iterative updating with short deformation, leading to a precise motion prediction for tasks such as landmark tracking. The iteration number within EMMA during training and inference is also examined for reaching a trade-off between performance and computational cost. Finally, we argue that, instead of learning the feature similarity directly, our proposed hybrid deformation can be injected within the tracking network using the proposed DPN for minimizing the tracking candidate searching space with a plausible deformation, which further increases the tracking performance.

\textbf{Complete v.s. Partial Deformation Decoder:} We first test different designs of the deformation learning network. As we have mentioned earlier, the deformation network takes the concatenated batch of the reference image, long and short interval images. The encoder extracts features and different decoder designs reconstruct the velocity field corresponding to different time intervals. A partial deformation network indicates that long and short deformation is learned from two independent decoders and a complete deformation network means that long and short deformation is learned from the shared decoder and only separated at the last convolution layer. From results shown in Table S2 Supplementary C, we observe that the complete design deformation network outperforms the partial design with less parameters, as the complete decoder benefits from learning the hybrid interval information not only in the encoder but also in the decoder. 

\textbf{EMMA Impacts:} We further test the effectiveness of EMMA to show the benefits of iteratively optimizing the long deformation based on short deformation as a latent motion variable.. From Table S2 Supplementary C, we show that EMMA can upgrade the deformation learning under both complete and partial motion decoder design. EMMA can support long short deformation to reduce the negatives of Jacobian shown in Fig. S1 Supplementary C. We further examine how different iteration numbers in complete motion E/M step during training and inference shown in Table S3. We can obtain a local optimum with a 5 iteration combination for reasonable performance without asking for further computational cost.

\textbf{Tracking with DPN:} Finally, we test how our proposed system can help achieve accurate landmark tracking with long/short deformation prior. With the feature map injected into the tracking network, we obtain lower tracking errors in both mean and standard deviation. Table S2 Supplementary C shows the quantitative result of accurate tracking from LSDM with DPN. Compared to the baseline tracking method without deformation modeling and DPN fusion, LSDM can directly learn the deformation between the exemplar and the follow-up instances. By combining priors through DPN, LSDM can minimize the search space for the tracklet and find the reliable candidate as a tracking object with plausible deformation, resulting in accurate tracking in spite of visual ambiguity.

\subsection{In-House Validation}
\label{sec:inhouse}
To validate the generalization of the proposed LSDM, we manually configure the training set for in-house testing. Specifically for CLUST2D, we select data samples from a specific hospital provider for testing, which is unseen for the network during the training stage. For AHNTU2D, we select data samples from a specific ultrasound modality for testing, which is unseen for the network during training on another ultrasound modality. The result is shown in Supplementary D. We aim to test whether or not a tracking method can generalize well to track the landmark within a video from the scanner it has never seen before. As reported in Tables S4 and S5 in Supplementary D, LSDM generalizes well on videos from the new ultrasound scanner during testing. LSDM our-performs the baseline SiamFC in both mean and std in tracking errors. By learning not only the feature similarity but also the deformation estimation, LSDM can handle the domain shift when testing on new scanners and new patients, showing high robustness and great potential for various clinical workflows such as radiation therapy. 

\subsection{Training Set Comparison}
\label{sec:trainingset}
LSDM achieves accurate tracking by learning an optimized deformation between the exemplar and follow-up instances, which can help the downstream tracking network for finding the best candidate with plausible deformation. As shown in Table \ref{Table:Training set result}, LSDM outperforms several baselines. Traditional correlation filtering based methods such as KCF and LCT cannot achieve satisfactory tracking performance on large-scale datasets due to various challenges. Even though methods like LCT incorporate historical information for updating tracking kernels, these baselines cannot produce accurate tracklets, lacking high-dimensional representation. Compared with tracking methods based on siamese networks such as SiamFC and SiamRPN, LSDM outperforms with lower tracking error in both mean and standard deviation. Specifically, LSDM outperforms SiamRPN with 1.61 lower in mean and 0.88 lower in standard deviation. Recall that SiamRPN contains an extra branch for minimizing the regression loss of the tracking object location directly. This indicates that only measuring the size (such as bounding box regression in SiamRPN) of the tracking object is not enough for accurate position estimation while an optimized deformation can be used as a prior for searching objects to improve tracking accuracy. We report an example of tracklet comparison in Fig. \ref{fig:trackresult} and additional examples in Fig. S3 Supplementary E. We can observe that our proposed LSDM can track the landmark accurately within long time ultrasound sequences while the other baselines fail in all different situations, e.g. distracted by other landmarks with a high visual similarity in Fig. \ref{fig:trackresult}, cannot handle the landmark deformation effectively resulted in accumulative tracking errors in Fig. S3 Supplementary E.

\subsection{Testset Result}
\label{sec:testset}
We report the group wise performance of LSDM on test sets in CLUST2D in Table \ref{table:grouptest} and total tracking error ranking in Table S6 Supplementary F. Compared with state-of-art methods and human labels, LSDM achieves competitive performance. Note that the 1st place method \cite{liu2020cascaded} uses heavy pre/processing techniques such as point detection and shadow removal. On the other hand, LSDM is a simple design but effective method without complex pre/preprocessing methods, while the learned deformation prior provides more insightful guiding for radiation therapy compared with other SOTA methods and no tracking, indicating the importance of accurate estimation of motion deformation during long time tracking. 

\subsection{Failure Case Analysis}
\label{sec:FailureCase}
We also report failure tracking examples caused by invalid field-of-view and lantency in system design. Both the failure case analysis and examples of failing tracking cases can be found in Supplementary G.

\section{Conclusion}
\label{sec:Conclusion}
In this paper, we proposed a multi-task based tracking method with a learnable deformation (LSDM). Instead of matching the feature similarity between exemplar and follow-up instances directly, we integrated the long-short diffeomorphism temporal learning as a deformation prior to search the candidate with the most plausible deformation to achieve accurate landmark tracking. LSDM took the full advantage of both long and short deformation by iteratively updating the estimation through the proposed expectation-maximization motion alignment (EMMA) module. Experiments on both public and private ultrasound landmark tracking datasets demonstrated the effectiveness and generalization of LSDM for clinical workflow. Compared with other competing methods, LSDM achieved superior tracking accuracy with a strong generalization capability across different scanners and modalities.

\ifCLASSOPTIONcaptionsoff
  \newpage
\fi



%



\bibliographystyle{IEEEtran}      
\bibliography{bare_jrnl} 

\clearpage
\onecolumn

\setcounter{table}{0}
\setcounter{figure}{0}
\setcounter{equation}{0}
\setcounter{page}{1}
\renewcommand\thefigure{S\arabic{figure}}
\renewcommand\thetable{S\arabic{table}}

\newtheorem*{theorem*}{Theorem}

\section{Supplementary A}

Table \ref{table:lr} summarizes the related works in  medical landmark tracking described in Sec. \Romannum{2}.

\begin{sidewaystable}
\caption{Summary of medical landmark tracking methods. In column \textbf{Learning}, \textquoteleft R\textquoteright \space means registration and \textquoteleft F\textquoteright \space means feature mapping. Column \textbf{DP} stands for deformation prior.}
\begin{adjustbox}{center}
\begin{tabular}{|c|c|c|c|c|c|l|}
\hline
Authors                             & Base Model                                                                                       & Learning               & Landmark                      & Modality                    & DP  & \multicolumn{1}{c|}{Remarks}                                                                                    \\ \hline
\multirow{2}{*}{Banerjee et al.\cite{banerjee2015combined}}   & \multirow{2}{*}{Grid Set}                                                                        & \multirow{2}{*}{R}     & \multirow{2}{*}{Liver}        & \multirow{2}{*}{US}         & \multirow{2}{*}{N} & Pro: Rigister to reference by tracking on two scale point set.                                                 \\
                                    &                                                                                                  &                        &                               &                             &                    & Con: Post-process on outlier rejection needed.                         \\ \hline
\multirow{2}{*}{Konig et al.\cite{konig2014non}}     & \multirow{2}{*}{Gradient Field}                                                       & \multirow{2}{*}{R}     & \multirow{2}{*}{Liver}        & \multirow{2}{*}{US}         & \multirow{2}{*}{Y} & Pro: Real time   tracking with gradient estimation on deformation.                                             \\
                                    &                                                                                                  &                        &                               &                             &                    & Con: Cannot   handle cummulative errors effectively.                                                           \\ \hline
\multirow{2}{*}{Yang et al.\cite{yang2011prediction}}      & \multirow{2}{*}{Shape Model}                                                            & \multirow{2}{*}{R}     & \multirow{2}{*}{Heart}        & \multirow{2}{*}{US}         & \multirow{2}{*}{Y} & Pro: Multi-View   multi-scale heart shape estimation for registration and tracking                             \\
                                    &                                                                                                  &                        &                               &                             &                    & Con: Not   end-to-end. Mannual fusion design involved.                                                         \\ \hline
\multirow{2}{*}{Bharadwaj et al.\cite{bharadwaj2021upgraded}} & \multirow{2}{*}{Kalman Filter}                                               & \multirow{2}{*}{F}     & \multirow{2}{*}{Liver}        & \multirow{2}{*}{US}         & \multirow{2}{*}{N} & Pro: Template update   strategy from Kalman Filter output.                                                     \\
                                    &                                                                                                  &                        &                               &                             &                    & Con: Cannot   estimate the landmark deformation. Unsatified tracking performance.                              \\ \hline
\multirow{2}{*}{Wu et al.\cite{wu2022fusion}}        & \multirow{2}{*}{Siamese Network}                                                                 & \multirow{2}{*}{F}     & \multirow{2}{*}{Liver}        & \multirow{2}{*}{US}         & \multirow{2}{*}{N} & Pro: Coarse-to-fine   training with drift correlation based on point distance                                  \\
                                    &                                                                                                  &                        &                               &                             &                    & Con: Cannot   handle deformed landmark with partial observation.                                               \\ \hline
\multirow{2}{*}{Liu et al.\cite{liu2020cascaded}}       & \multirow{2}{*}{Siamese Network}                                                                 & \multirow{2}{*}{F}     & \multirow{2}{*}{Liver}        & \multirow{2}{*}{US}         & \multirow{2}{*}{N} & Pro: Multi-scale   tracking network.                                                                           \\
                                    &                                                                                                  &                        &                               &                             &                    & Con:   Training performance relies on the quality of generated landmark points.\\ \hline
\multirow{2}{*}{Cifor et al.\cite{cifor2013hybrid}}     & \multirow{2}{*}{Shape Model Ensemble}                                                            & \multirow{2}{*}{R}     & \multirow{2}{*}{Liver}        & \multirow{2}{*}{US}         & \multirow{2}{*}{Y} & Pro: Registration   based tracking using ensemble deformation models.                                          \\
                                    &                                                                                                  &                        &                               &                             &                    & Con:Lacking   pricise template matching design.                                                                \\ \hline
\multirow{2}{*}{Royer et al.\cite{royer2017real}}     & \multirow{2}{*}{Mechanical Simulation}                                                           & \multirow{2}{*}{R}     & \multirow{2}{*}{Liver}        & \multirow{2}{*}{US}         & \multirow{2}{*}{N} & Pro: Vertex position   estimation using visual information and machenical simulation.                          \\
                                    &                                                                                                  &                        &                               &                             &                    & Con: Heavy inference workload during long time series   images.           \\ \hline
\multirow{2}{*}{Gomariz et al.\cite{gomariz2019siamese}}   & \multirow{2}{*}{Siamese Network}                                                                 & \multirow{2}{*}{F}     & \multirow{2}{*}{Liver}        & \multirow{2}{*}{US}         & \multirow{2}{*}{N} & Pro: Using previous   location as a localization prior.                                                        \\
                                    &                                                                                                  &                        &                               &                             &                    & Con: Lacking   deformation estimation.                                                                         \\ \hline
\multirow{2}{*}{Makhinya et al.\cite{makhinya2015motion}}  & \multirow{2}{*}{Optical Flow}                                                                    & \multirow{2}{*}{R}     & \multirow{2}{*}{Liver}        & \multirow{2}{*}{US}         & \multirow{2}{*}{Y} & Pro: Optical flow   based motion estimation.                                                                   \\
                                    &                                                                                                  &                        &                               &                             &                    & Con: Manual   designed vessel features. Lacking template matching design.                                      \\ \hline
\multirow{2}{*}{Shepard et al.\cite{shepard2017block}}   & \multirow{2}{*}{Block Matching}                                                                  & \multirow{2}{*}{F}     & \multirow{2}{*}{Liver}        & \multirow{2}{*}{US}         & \multirow{2}{*}{N} & Pro: Multi-scale   block matching method.                                                                      \\
                                    &                                                                                                  &                        &                               &                             &                    & Con:   Performance relies on local block quality. Lacking deformation estimation.                              \\ \hline
\multirow{2}{*}{Ye et al.\cite{ye2021deeptag}}        & \multirow{2}{*}{Motion Tracking}                                                                 & \multirow{2}{*}{R}     & \multirow{2}{*}{Heart}        & \multirow{2}{*}{Tagged MRI} & \multirow{2}{*}{Y} & Pro:   Forward-Backward motion modeling.                                                                       \\
                                    &                                                                                                  &                        &                               &                             &                    & Con: No   explicit discriminal learning on landmarks.                                                          \\ \hline
\multirow{2}{*}{Qin et al. \cite{qin2020biomechanics}}         & \multirow{2}{*}{Motion Tracking}                                                                 & \multirow{2}{*}{R}     & \multirow{2}{*}{Heart}        & \multirow{2}{*}{MRI}        & \multirow{2}{*}{Y} & Pro:   Biomechanics-informed motion modeling                                                                   \\
                                    &                                                                                                  &                        &                               &                             &                    & Con: No   historical information during motion modeling.                                                       \\ \hline
\multirow{2}{*}{Rangamani et al.\cite{rangamani2016landmark}} & \multirow{2}{*}{CNN+RNN}                                                                         & \multirow{2}{*}{F}     & \multirow{2}{*}{Liver}        & \multirow{2}{*}{US}         & \multirow{2}{*}{N} & Pro: RNN location   predictor based on CNN features.                                                           \\
                                    &                                                                                                  &                        &                               &                             &                    & Con:Heavy   weight network design. No deformation learning on landmarks.                                       \\ \hline
\multirow{2}{*}{Huang et al.\cite{huang2019attention}}     & \multirow{2}{*}{CNN+LSTM}                                                                        & \multirow{2}{*}{F}     & \multirow{2}{*}{Liver}        & \multirow{2}{*}{US}         & \multirow{2}{*}{N} & Pro: LSTM for   location refinement.                                                                           \\
                                    &                                                                                                  &                        &                               &                             &                    & Con: Lacking interpretation during refinement process.                          \\ \hline
\multirow{2}{*}{Ha et al.\cite{ha2018model}}        & \multirow{2}{*}{Motion Tracking}                                                                 & \multirow{2}{*}{R}     & \multirow{2}{*}{Abdomen}      & \multirow{2}{*}{4D MRI}     & \multirow{2}{*}{Y} & Pro: Coupled conves   optimization for real time motion estimation.                                            \\
                                    &                                                                                                  &                        &                               &                             &                    & Con:   Landmark tracking performance relies on the choice of block template.                                   \\ \hline
\multirow{2}{*}{Shen et al.\cite{shen2018online}}      & \multirow{2}{*}{KCF}                                                                             & \multirow{2}{*}{F}     & \multirow{2}{*}{Liver}        & \multirow{2}{*}{US}         & \multirow{2}{*}{N} & Pro: Optimized KCF   for real time landmark tracking.                                                          \\
                                    &                                                                                                  &                        &                               &                             &                    & Con:   Unsatified performance. Lacking deformation learning on landmarks.                                      \\ \hline
\multirow{2}{*}{Wilms et al.\cite{wilms2016model}}     & \multirow{2}{*}{Block Matching}                                                                  & \multirow{2}{*}{R}     & \multirow{2}{*}{Adbomen}      & \multirow{2}{*}{4D MRI}     & \multirow{2}{*}{Y} & Pro: Coarse-to-fine   training with model based regularization.                                                \\
                                    &                                                                                                  &                        &                               &                             &                    & Con:   Performance relies on block matching.                                                                   \\ \hline
\multirow{2}{*}{Williamson et al.\cite{williamson2018ultrasound}}  & \multirow{2}{*}{Distance Modeling}                                                               & \multirow{2}{*}{F}     & \multirow{2}{*}{Liver}        & \multirow{2}{*}{US}         & \multirow{2}{*}{N} & Pro: Multi-template   based aggregation.                                                                       \\
                                    &                                                                                                  &                        &                               &                             &                    & Con:   Performance relies on keypoint selection.                                                               \\ \hline
\multirow{2}{*}{Ours}               & \multirow{2}{*}{\begin{tabular}[c]{@{}c@{}}Motion aware \\ block matching  network\end{tabular}} & \multirow{2}{*}{F + R} & \multirow{2}{*}{Liver/Kidney} & \multirow{2}{*}{US}         & \multirow{2}{*}{Y} & Pro: Hybrid motion   modeling for landmark deformation matching.                                               \\
                                    &                                                                                                  &                        &                               &                             &                    & Con: No   specific strategy handling poor quality image with limited field-of-view.                            \\ \hline
\end{tabular}
\label{table:lr}
\end{adjustbox}
\end{sidewaystable}

\section{Supplementary B}
Below we present the proof of Theorem III.1. We first restate Theorem III.1.
\begin{theorem*}
\label{the:emma convergence}
The expectation maximization motion alignment can be converged with an optimum $\theta^{\ast}$ within the iterations of $N$ times where:
\begin{equation}
    \log p(\phi_l | \phi_s, \theta^{\ast}) \geq \log p (\phi_l | \phi_s, \theta^{n}), n = 1, …, N, N \rightarrow \infty
\end{equation}
\end{theorem*}

\begin{proof}
Following standard Expectation Maximization, we have:
\begin{equation}
\label{eqn:standard em}
    \log p(\phi_l | \theta) = \log \left \{ \frac{p(\phi_l, \phi_s | \theta)}{p(\phi_s | \phi_l, \theta)} \right \}
\end{equation}

by taking the expectation on both sides of Eq.(\ref{eqn:standard em}), w.r.t $p(\phi_s | \phi_l, \theta^{n})$ at iteration $n$, the left part of Eq.(\ref{eqn:standard em}) equals to:
\begin{equation}
\label{eqn:left}
\begin{split}
        L &= \int_{\phi_s} p(\phi_s | \phi_l, \theta^{n}) \log p(\phi_l | \theta) d \phi_s\\
	&= \log p(\phi_l | \theta)
\end{split}
\end{equation}
The right part of Eq.(\ref{eqn:standard em}) equals to:
\begin{equation}
\label{eqn:right}
\begin{split}
        R = & \int_{\phi_s} p(\phi_s|\phi_l, \theta^{n}) \log p(\phi_l, \phi_s | \theta) d\phi_s - \int_{\phi_s}p(\phi_s|\phi_l, \theta^{n}) \log p (\phi_s|\phi_l, \theta) d \phi_s
\end{split}
\end{equation}
For the second term in Eq.(\ref{eqn:right}), we have:

\begin{equation}
\begin{split}
\label{eqn:convergence}
R &=\int_{\phi_s}p(\phi_s|\phi_l, \theta^{n}) \log \left \{ \frac{p(\phi_s|\phi_l, \theta^{n+1})}{p(\phi_s|\phi_l, \theta^{n})} \right \}\\
&\leq \log \int_{\phi_s}p(\phi_s|\phi_l, \theta^{n})\frac{p(\phi_s|\phi_l, \theta^{n+1})}{p(\phi_s|\phi_l, \theta^{n})}d\phi_s\\
&\leq \log \int_{\phi_s}p(\phi_s|\phi_l, \theta^{n+1})d\phi_s\\
&\leq \log \left (1 \right )\\ 
&\leq 0
\end{split}
\end{equation}
\end{proof}

By combining Eqs.(\ref{eqn:left}) and (\ref{eqn:right}), we have the log-likelihood of long deformation given parameter $\theta$:
\begin{equation}
\label{eqn:loglikelihood}
\begin{split}
    \log p(\phi_l|\theta) &= \underbrace{E_{q(\phi_s)}\left [\ \log\frac{p(\phi_l, \phi_s | \theta)}{q(\phi_s)}\right ]\ }_\text{Motion ELBO} + \underbrace{\int q(\phi_s) \log \left \{ \frac{q(\phi_s)}{p(\phi_l|\phi_s, \theta)} \right \}d\phi_s}_\text{KL divergence}\\
\end{split}
\end{equation}

Following the structural definition in Eq.(\ref{eqn:loglikelihood}), EMMA first starts with randomly initialized parameter $\theta$ and construct the expectation expression at iteration $n$:
\begin{equation}
\begin{split}
    \text{E-step:}&\\
    &p(\phi_s|\phi_l, \theta^{n}) \rightarrow E_{\phi_s|\phi_l, \theta^{t}}\left [\ \log p(\phi_l, \phi_s|\theta) \right ]\
\end{split}
\end{equation}
and update the parameter for maximizing the constructed expectation:
\begin{equation}
\begin{split}
    \text{M-step:}&\\
    &\theta^{n+1} = \argmaxA_{\theta} E_{\phi_s|\phi_l, \theta^{n}} \left [\ \log p(\phi_s, \phi_l|\theta) \right ]\
\end{split}
\end{equation}
The ideal physical meaning of EMMA indicates that, once $\theta$ is converged to $\theta^{\ast}$, the approximate distribution $q(\phi_s)$ is equal to the true posterior $p(\phi_s|\phi_l, \theta^{\ast})$. The KL divergence in Eq.(\ref{eqn:loglikelihood}) approaches zero and the ELBO is bounded by the maximum. Thus, the log-likelihood of long deformation is maximized, meaning EMMA can output the most plausible long deformation.

\newpage
\section{Supplementary C: Ablation Study}
\begin{figure*}[thp]
    \centering
    \includegraphics[width=16cm, keepaspectratio]{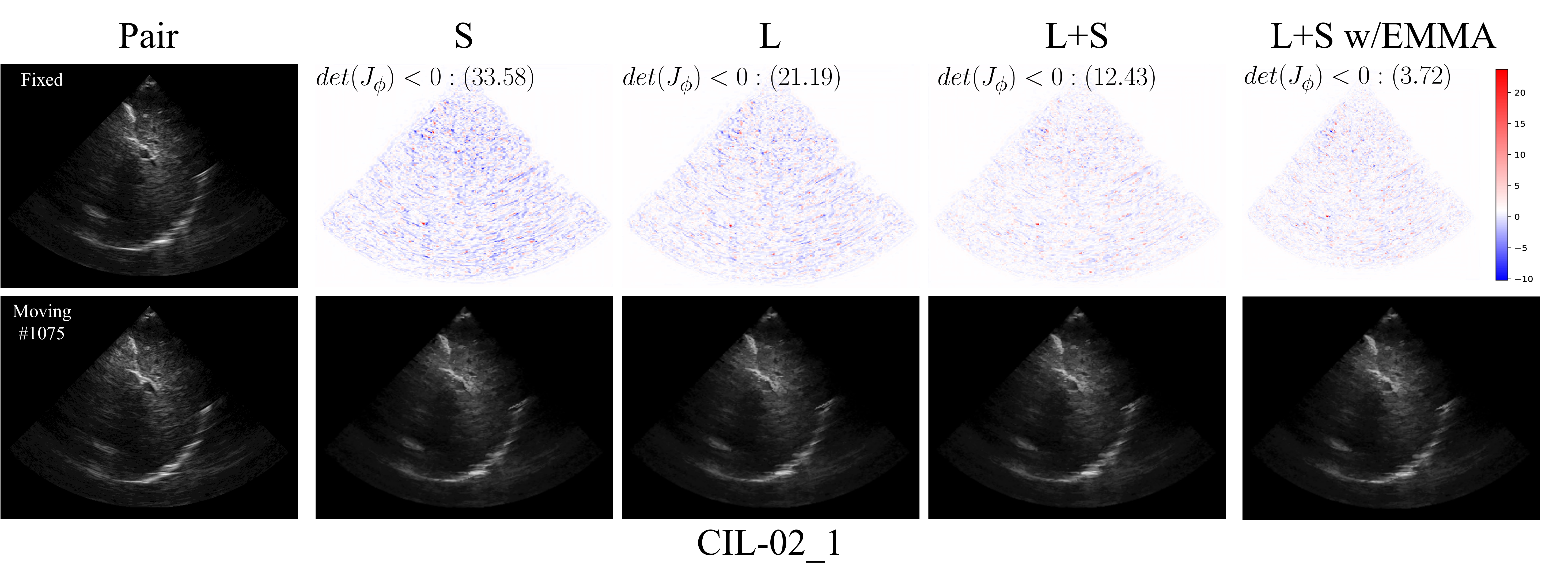}
    \caption{Qualitative deformation comparison of the proposed LSDM based on different deformation module combinations. We select one representative dataset entriy to demonstrate superior LSDM deformation performance. For each entry, we show the fixed template at frame 0 (top-left), selected target frame (bottom-left), warped images based on different deformation combination (bottom row) and visualization of determinant of jacobian matrix from a different displacement field (top row), where red indicates the determinant of jacobian is greater than 1 and blue indicates the value of the determinant of jacobian is negative. We can observe a progressive smooth setting with proposed long-short deformation and EMMA module. Together with DPN, the LSDM learns to generate the optimal deformation for downstream tasks like tracking. Best viewed in color.}
\label{fig:morphresult}
\end{figure*}

In this section, we show the supplemental ablation study results of LSDM, analyzed in Sec. V. Table \ref{Table:Ablation Study} shows the component ablation study results tested on the CLUST2D training set. Table \ref{Table:EMMA Iter} tests different impacts of different EM iteration numbers. Fig. \ref{fig:morphresult} shows the qualitative deformation comparison of LSDM based on different deformation module combinations.  

\begin{table*}[h]
\caption{Quantitative results of ablation studies on LSDM regarding complete v.s partial deformation prior network, w/o EMMA and feature fusion w/o deformation pyramid network.}
\small
\centering\centering
\begin{tabular}{|l|llll|l|}
\hline
\multicolumn{1}{|c|}{\multirow{2}{*}{}} & \multicolumn{4}{c|}{Components}                                                                & Metrics         \\ \cline{2-6} 
\multicolumn{1}{|c|}{}                  & \multicolumn{1}{l|}{Complete} & \multicolumn{1}{l|}{Partial} & \multicolumn{1}{l|}{EMMA} & DPN & TE Mean +/- Std \\ \hline
\multirow{2}{*}{Deformation Prior}    & \multicolumn{1}{l|}{\ding{51}}        & \multicolumn{1}{l|}{}        & \multicolumn{1}{l|}{}     &     & 2.63 +/- 2.11   \\ \cline{2-6} 
                                        & \multicolumn{1}{l|}{}         & \multicolumn{1}{l|}{\ding{51}}       & \multicolumn{1}{l|}{}     &     & 2.69 +/- 2.87   \\ \hline
\multirow{2}{*}{EMMA}                   & \multicolumn{1}{l|}{\ding{51}}        & \multicolumn{1}{l|}{}        & \multicolumn{1}{l|}{\ding{51}}    &     & 1.21 +/- 2.19   \\ \cline{2-6} 
                                        & \multicolumn{1}{l|}{}         & \multicolumn{1}{l|}{\ding{51}}       & \multicolumn{1}{l|}{\ding{51}}    &     & 1.56 +/- 1.73   \\ \hline
\multirow{2}{*}{DPN}                    & \multicolumn{1}{l|}{\ding{51}}        & \multicolumn{1}{l|}{}        & \multicolumn{1}{l|}{\ding{51}}    & \ding{51}   & 0.92 +/- 0.76   \\ \cline{2-6} 
                                        & \multicolumn{1}{l|}{}         & \multicolumn{1}{l|}{\ding{51}}       & \multicolumn{1}{l|}{\ding{51}}    & \ding{51}   & 0.81 +/- 0.98   \\ \hline
\end{tabular}
\label{Table:Ablation Study}
\end{table*}

\begin{table*}[hbt]
\caption{Quantitative results of EMMA iteration number test on CLUST2D Training set.}
\small
\centering\centering
\begin{tabular}{|l|l|}
\hline
\# EMMA   Iteration & TE Mean +/- Std \\ \hline
1                   & 1.46 +/- 1.72   \\ \hline
5                  & 0.81 +/- 0.98   \\ \hline
10                  & 0.93 +/- 1.37   \\ \hline

\end{tabular}
\label{Table:EMMA Iter}
\end{table*}

\newpage
\section{Supplementary D: In House Test}

\begin{figure*}[thp]
    \centering
    \includegraphics[width=\textwidth, keepaspectratio]{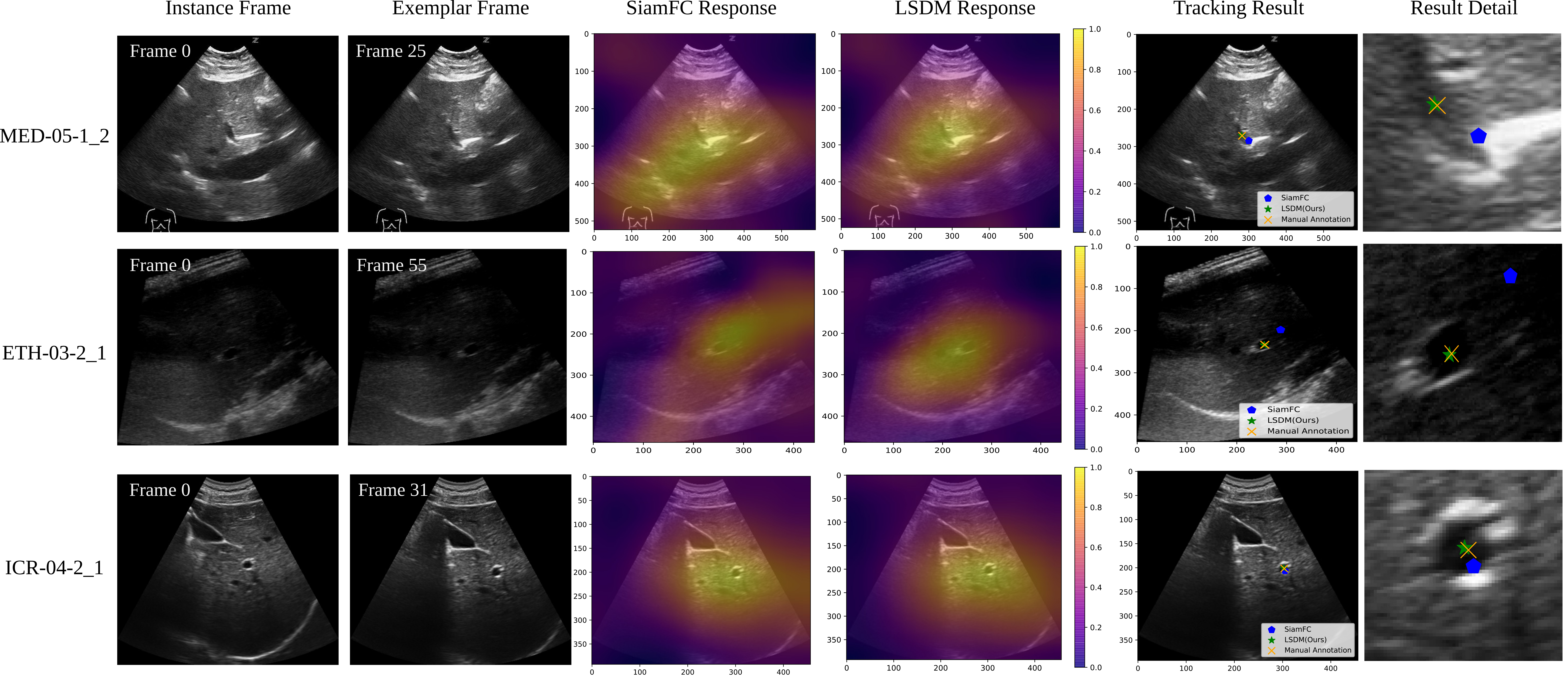}
    \caption{Qualitative landmark tracking response comparison between the baseline model (SiamFC) and our proposed LSDM on the CLUST2D training set. From left to right, each column represents the instance frame (frame 0), selected exemplar frame, SiamFC response, LSDM response, tracking result visualization and zoomed-in patch. Best viewed in color.}
\label{fig:response}
\end{figure*}

In this section, we show the supplemental in-house test results of LSDM, analyzed in Sec. V-B. Table \ref{Table:CLUST2D-InHouseTest} shows the in-house test result of LSDM and SiamFC \cite{bertinetto2016fully} on the CLUST2D training set, where the test partition is never shown during the training process. Table \ref{Table:AHNTU2D-InHouseTest}  shows the in-house test result of LSDM and SiamFC \cite{bertinetto2016fully} on the CLUST2D training set, where the test modality is never shown during the training process. Fig. \ref{fig:response} shows the qualitative response comparison between LSDM and SiamFC during the in-house test.

\begin{table*}[h]
\caption{Quantitative result comparison between LSDM and the baseline on the CLUST2D Training set with in-house validation setting, with respect to mean, standard deviation and 95th tracking error.}
\small
\centering\centering
\begin{tabular}{|c|cc|cc|cc|c|}
\hline
\multirow{2}{*}{In-House Partition} & \multicolumn{2}{c|}{Mean}          & \multicolumn{2}{c|}{Std}           & \multicolumn{2}{c|}{95th}          & \multirow{2}{*}{Scanner Type} \\ \cline{2-7}
                                    & \multicolumn{1}{c|}{SiamFC} & LSDM & \multicolumn{1}{c|}{SiamFC} & LSDM & \multicolumn{1}{c|}{SiamFC} & LSDM &                               \\ \hline
CIL                                 & \multicolumn{1}{c|}{2.01}   & 1.82 & \multicolumn{1}{c|}{3.47}   & 1.63 & \multicolumn{1}{c|}{11.49}  & 3.81 & Ultrasonix MDP                \\ \hline
ETH                                 & \multicolumn{1}{c|}{5.33}   & 1.98 & \multicolumn{1}{c|}{10.16}  & 1.21 & \multicolumn{1}{c|}{17.3}   & 4.67 & Siemens Antares               \\ \hline
ICR                                 & \multicolumn{1}{c|}{1.09}   & 2.19 & \multicolumn{1}{c|}{3.22}   & 1.76 & \multicolumn{1}{c|}{5.64}   & 3.76 & Elekta   Clarity-Ultrasonix   \\ \hline
MED1                                & \multicolumn{1}{c|}{3.17}   & 1.35 & \multicolumn{1}{c|}{2.46}   & 1.9  & \multicolumn{1}{c|}{7.71}   & 2.91 & Zonare z.one                  \\ \hline
MED2                                & \multicolumn{1}{c|}{4.93}   & 3.19 & \multicolumn{1}{c|}{9.27}   & 1.31 & \multicolumn{1}{c|}{19.15}  & 5.18 & DiPhAs Fraunhofer             \\ \hline
\end{tabular}
\label{Table:CLUST2D-InHouseTest}
\end{table*}

\begin{table*}[h]
\caption{Quantitative result comparison between LSDM and the baseline on the AHNTU2D Training set with in-house validation setting on different modalities, with respect to mean, standard deviation and 95th tracking error.}
\small
\centering\centering
\begin{tabular}{|l|lll|lll|}
\hline
\multirow{2}{*}{AHNTU} & \multicolumn{3}{l|}{AHNTU-N}                                   & \multicolumn{3}{l|}{AHNTU-C}                                   \\ \cline{2-7} 
                       & \multicolumn{1}{l|}{Mean} & \multicolumn{1}{l|}{Std}  & TE95th & \multicolumn{1}{l|}{Mean} & \multicolumn{1}{l|}{Std}  & TE95th \\ \hline
SiamFC \cite{bertinetto2016fully}                 & \multicolumn{1}{l|}{2.74} & \multicolumn{1}{l|}{3.08} & 6.24   & \multicolumn{1}{l|}{4.41} & \multicolumn{1}{l|}{6.77} & 13.17  \\ \hline
LSDM(Ours)             & \multicolumn{1}{l|}{1.31} & \multicolumn{1}{l|}{1.76} & 3.72   & \multicolumn{1}{l|}{2.39} & \multicolumn{1}{l|}{4.12} & 10.83  \\ \hline
\end{tabular}
\label{Table:AHNTU2D-InHouseTest}
\end{table*}

\newpage
\section{Supplementary E: Training Set Comparison}

\begin{figure*}[thp]
    \centering
    \includegraphics[width=\textwidth, keepaspectratio]{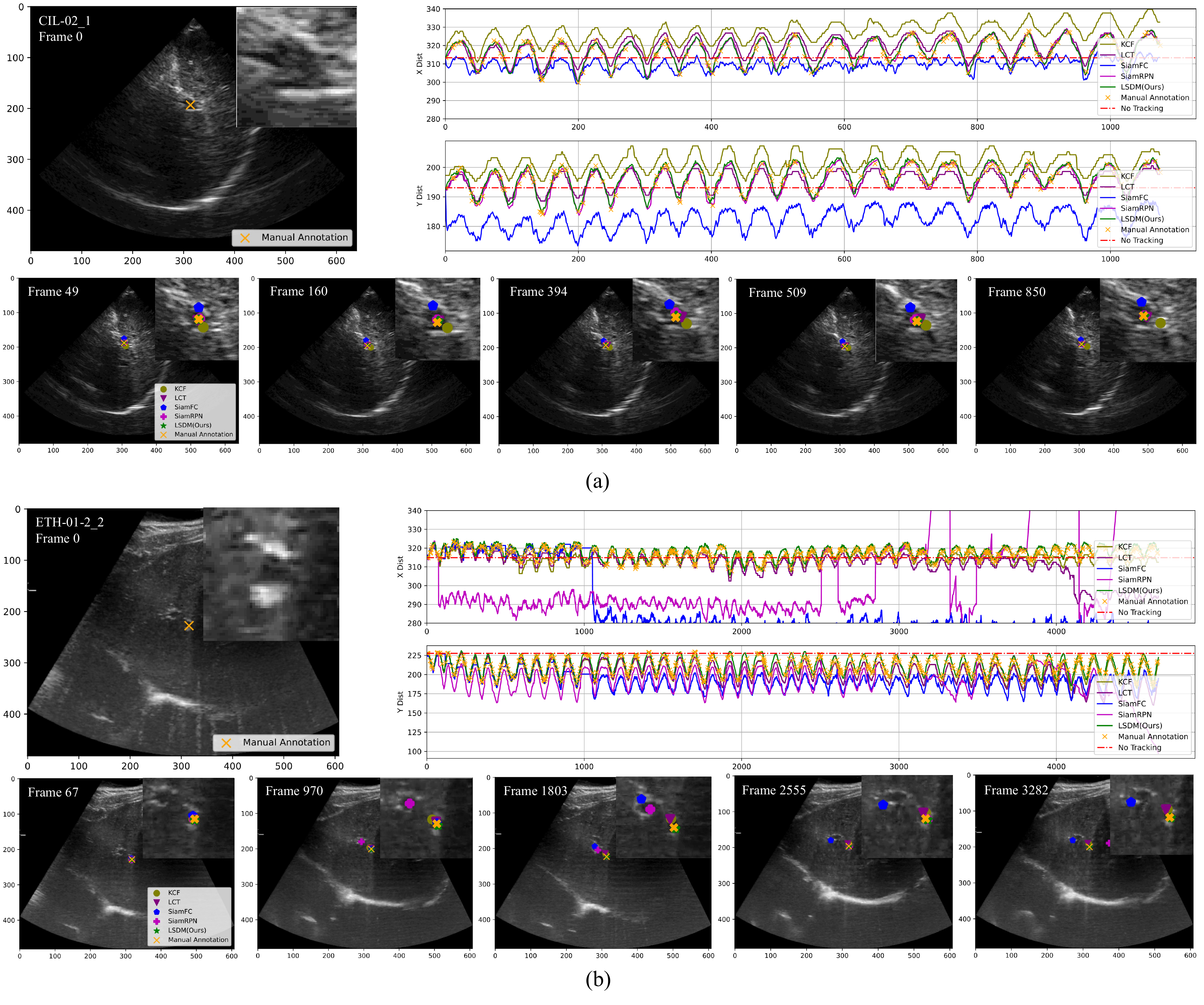}
    \caption{Additional qualitative comparison of different baseline models and the proposed LSDM by tracking validation on the CLUST2D training set. We select other two representative dataset entries to demonstrate LSDM superior tracking performance. We show the start frame at frame 0 (top-left), landmark center coordinate tracklet comparison (top-right) and selected frames with landmark tracking results (bottom). Best viewed in color.}
\label{fig:tracklet}
\end{figure*}

Fig. \ref{fig:tracklet} shows additional qualitative tracklet comparision between different baseline models and our proposed LSDM tested on CLUST2D training set, described in Sec. \Romannum{5}-C.

\newpage
\section{Supplementary F: Test Set Ranking Result}

Table S6 summarizes the comparisons of tracking errors evaluated on the CLUST2D and AHNTU test sets. We observe that our proposed framework achieves 
superior or competitive tracking performance against other state-of-the-art medical landmark tracking methods. Note that LSDM is a simeple but effective multi-task design without complicated pre-/post-processings. LSDM achieves with more stable performance during various tests, including gourp-wise test set result described in Sec. \Romannum{5}-D.

\begin{table}[tbh]
\caption{Quantitative overall tracking performance comparison of LSDM against other state-of-the-art methods on CLUST2D and AHNTU Test set.}
\small
\centering\centering
\begin{tabular}{|l|ccc|}
\hline
\multicolumn{1}{|c|}{\multirow{2}{*}{CLUST2D}} & \multicolumn{3}{c|}{Overall}                                     \\ \cline{2-4} 
\multicolumn{1}{|c|}{}                         & \multicolumn{1}{c|}{Mean}  & \multicolumn{1}{c|}{Std}   & TE95th \\ \hline
Liu et al. \cite{liu2020cascaded}                                     & \multicolumn{1}{c|}{0.69}  & \multicolumn{1}{c|}{0.67}  & 1.57   \\ \hline
Williamson et   al. \cite{williamson2018ultrasound}                            & \multicolumn{1}{c|}{0.74}  & \multicolumn{1}{c|}{1.03}  & 1.85   \\ \hline
Wu et al. \cite{wu2022fusion}                                     & \multicolumn{1}{c|}{0.8}   & \multicolumn{1}{c|}{1.16}  & 2.29   \\ \hline
\rowcolor{lavender}
LSDM (Ours)                                    & \multicolumn{1}{c|}{1.01}  & \multicolumn{1}{c|}{1.16}  & 2.21   \\ \hline
Shen et al. \cite{shen2019discriminative}                                   & \multicolumn{1}{c|}{1.11}  & \multicolumn{1}{c|}{0.91}  & 2.68   \\ \hline
Hallack et al. \cite{hallack2015robust}                                 & \multicolumn{1}{c|}{1.21}  & \multicolumn{1}{c|}{3.17}  & 2.82   \\ \hline
Gomariz et al. \cite{gomariz2019siamese}                                 & \multicolumn{1}{c|}{1.34}  & \multicolumn{1}{c|}{2.57}  & 2.95   \\ \hline
Makhinya and   Golsel \cite{makhinya2015motion}                          & \multicolumn{1}{c|}{1.44}  & \multicolumn{1}{c|}{2.8}   & 3.62   \\ \hline
Bharadwaj S.,   et al. \cite{bharadwaj2021upgraded}                        & \multicolumn{1}{c|}{1.60}   & \multicolumn{1}{c|}{3.69}  & 4.21   \\ \hline
Kondo \cite{kondo2015liver}                                          & \multicolumn{1}{c|}{2.91}  & \multicolumn{1}{c|}{10.52} & 5.18   \\ \hline
Nouri D. \&   Rothberg A. \cite{nouri2015liver}                     & \multicolumn{1}{c|}{3.35}  & \multicolumn{1}{c|}{5.21}  & 14.19  \\ \hline
No Tracking                                    & \multicolumn{1}{c|}{6.45}  & \multicolumn{1}{c|}{5.11}  & 16.48  \\ \hline
\multicolumn{1}{|c|}{\multirow{2}{*}{AHNTU}}   & \multicolumn{3}{c|}{Overall}                                     \\ \cline{2-4} 
\multicolumn{1}{|c|}{}                         & \multicolumn{1}{c|}{Mean}  & \multicolumn{1}{c|}{Std}   & TE95th \\ \hline
\rowcolor{lavender}
LSDM (Ours)                                    & \multicolumn{1}{c|}{1.80}   & \multicolumn{1}{c|}{2.33}  & 4.17   \\ \hline
SiamFC \cite{bertinetto2016fully}                                         & \multicolumn{1}{c|}{4.11}  & \multicolumn{1}{c|}{5.64}  & 10.49  \\ \hline
No Tracking                                    & \multicolumn{1}{c|}{13.71} & \multicolumn{1}{c|}{11.6}  & 31.74  \\ \hline
\end{tabular}
\label{table:SOTA}
\end{table}

\newpage
\section{Supplementary G: Failure Case Analysis}

\begin{figure*}[thp]
    \centering
    \includegraphics[width=\textwidth, keepaspectratio]{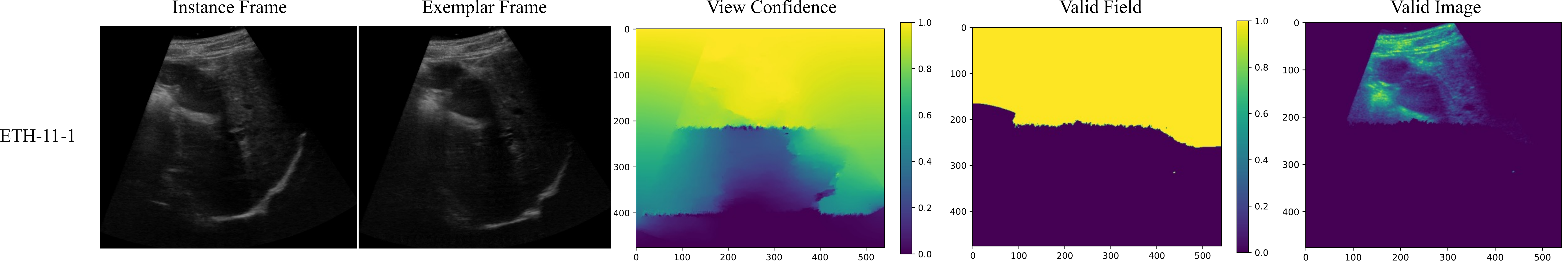}
    \caption{Ultrasound image field-of-view confidence score visualization using \cite{karamalis2012ultrasound}. We select a representative case from the CLUST2D test set. From left to right: instance frame at frame 0, selected exemplar frame, generated field-of-view confidence score, valid field-of-view based on score threshold and high visible image area within the valid field-of-view. As the landmark enters the invalid field-of-view, it is less visible even to human expert eyes, decreasing the tracking performance to shadow agnostic methods like LSDM. Best viewed in color.}
\label{fig:fieldofviewscore}
\end{figure*}

We report the examples of failed tracking cases with detailed analysis mentioned in Sec. \Romannum{5}-E. Compared with other high-signal-capacity modalities such as CT and MRI, Ultrasound imaging is less competent to handle the cases with low signal-to-noise ratios. It is very difficult to extract local features for accurate tracking from such highly noisy environments. In Fig. \ref{fig:fieldofviewscore}, we show a typical failure case, where the valid visible area of the input image is very narrow and a large part of the invalid view is the shadow area caused by insufficient ultrasound gel. It is less visible even to human experts when the landmark moves into the shadow area. The effect of LSDM on landmark matching and deformation estimation is limited, resulting in inaccurate tracking performance. In the future work, we will extend LSDM and take the advantage of the shadow area segmentation module as proposed in \cite{karamalis2012ultrasound}, \cite{meng2019weakly}, let LSDM be self-adaptive to the shadow areas.

\begin{figure}[thp]
    \centering
    \includegraphics[width=0.35\textwidth, keepaspectratio]{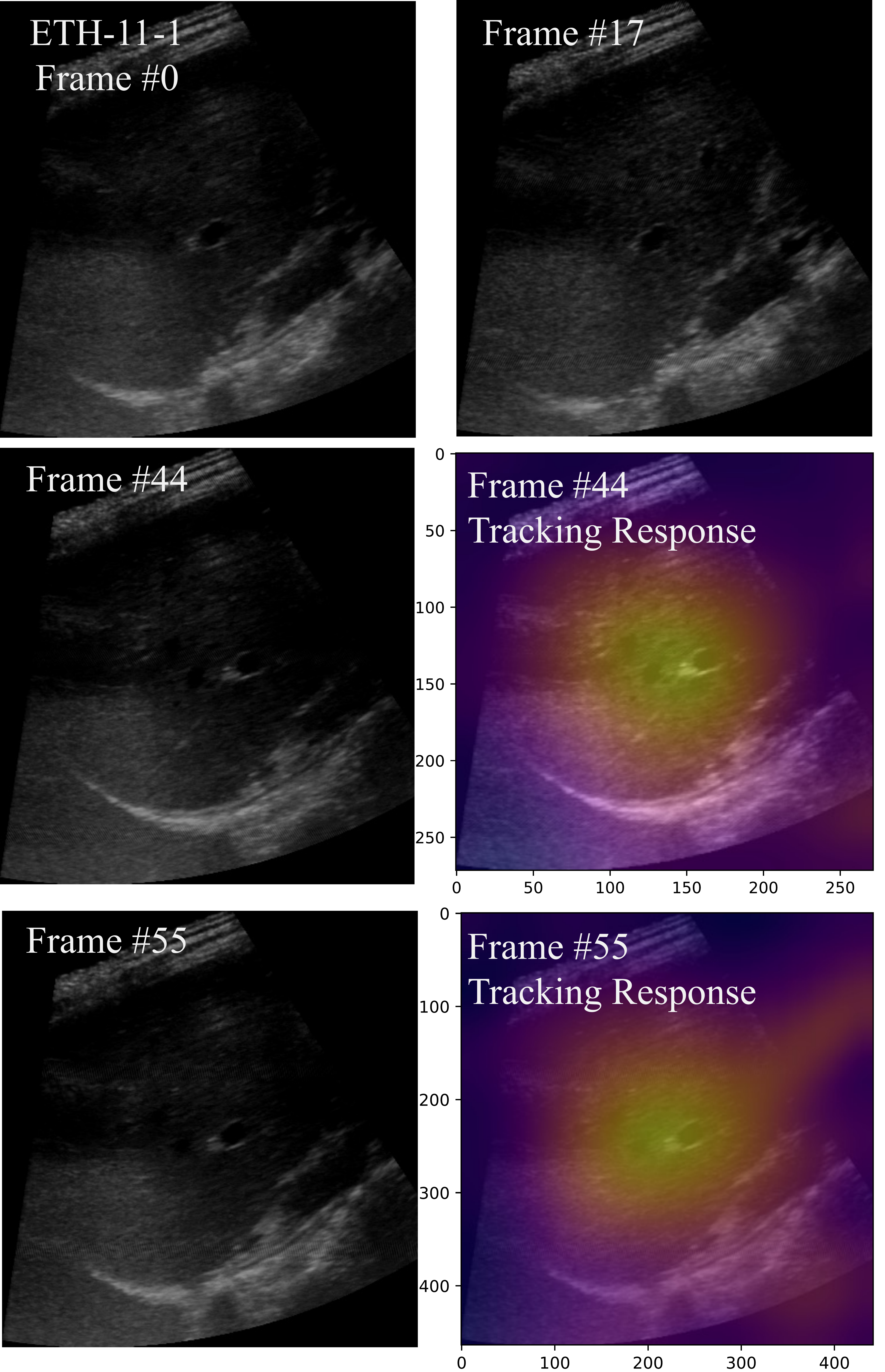}
    \caption{Visualization of tracking latency. We select a representative case from the CLUST2D training set. From top-left to bottom-right: instance frame at frame 0, mid-interval frame at frame 17, one exemplar frame at frame 44, LSDM tracking response on frame 44, one exemplar frame at frame 55 and LSDM tracking response on frame 55. We can observe a tracking latency within online tracking methods. Best viewed in color.}
\label{fig:latency}
\end{figure}

Although LSDM has achieved stable and highly accurate results, we discover another type of factors affecting the tracking performance, namely latency matching. This is because we built LSDM on the online tracking system. During the training and inference stage, LSDM can only act on the current t-th frame and agnostic on frames after the $t$-th frame. This leads to online latency in the estimation of the landmark by LSDM when the landmark changes in a nonlinear acceleration (shown in Fig. \ref{fig:latency}). In the follow-up deployment, we will adapt the temporal receptive field of LSDM in both online and offline modes so that LSDM can access the image sequences after time t to minimize the latency \cite{yao2016part}, \cite{dong2020clnet}, \cite{yan2021alpha}.

%








\end{document}